\documentclass[sn-basic]{sn-jnl}

\usepackage[utf8]{inputenc}%
\usepackage{graphicx}%
\usepackage{multirow}%
\usepackage{amsmath,amssymb,amsfonts}%
\usepackage{amsthm}%
\usepackage{mathrsfs}%
\usepackage[title]{appendix}%
\usepackage{xcolor}%
\usepackage{textcomp}%
\usepackage{manyfoot}%
\usepackage{booktabs}%
\usepackage{algorithm}%
\usepackage{algorithmicx}%
\usepackage{algpseudocode}%
\usepackage{listings}
\usepackage{tikz}

\theoremstyle{thmstyleone}%
\theoremstyle{thmstyletwo}%

\theoremstyle{thmstylethree}%

\begin{document}

\title[Article Title]{Exploring the Noise Resilience of Successor Features and Predecessor Features Algorithms in One and Two-Dimensional Environments}


\author*{\fnm{Hyunsu} \sur{Lee}}\email{hyunsu.lee@pusan.ac.kr}

\affil*{\orgdiv{Department of Physiology}, \orgname{School of Medicine, Pusan National University}, \orgaddress{\street{Busandaehak-ro}, \city{Yangsan}, \postcode{50612},  \country{Republic of Korea}}}

\abstract{Based on the predictive map theory of spatial learning in animals, this study delves into the dynamics of Successor Feature (SF) and Predecessor Feature (PF) algorithms within noisy environments. Utilizing Q-learning and Q($\lambda$) learning as benchmarks for comparative analysis, our investigation yielded unexpected outcomes. Contrary to prevailing expectations and previous literature where PF demonstrated superior performance, our findings reveal that in noisy environments, PF did not surpass SF. In a one-dimensional grid world, SF exhibited superior adaptability, maintaining robust performance across varying noise levels. This trend of diminishing performance with increasing noise was consistent across all examined algorithms, indicating a linear degradation pattern.
The scenario shifted in a two-dimensional grid world, where the impact of noise on algorithm performance demonstrated a non-linear relationship, influenced by the $\lambda$ parameter of the eligibility trace. This complexity suggests that the interaction between noise and algorithm efficacy is tied to the environmental dimensionality and specific algorithmic parameters.
Furthermore, this research contributes to the bridging discourse between computational neuroscience and reinforcement learning (RL), exploring the neurobiological parallels of SF and PF learning in spatial navigation. Despite the unforeseen performance trends, the findings enrich our comprehension of the strengths and weaknesses inherent in RL algorithms. This knowledge is pivotal for advancing applications in robotics, gaming AI, and autonomous vehicle navigation, underscoring the imperative for continued exploration into how RL algorithms process and learn from noisy inputs.}




\maketitle

\section{Introduction}\label{sec:intro}

Reinforcement learning (RL) has made significant progress in fields such as robotics and autonomous vehicles, with breakthrough technologies such as AlphaZero \citep{schrittwieser2020,silver2017,kaufmann2023,wurman2022,ghadirzadeh2021,smith2023}. This computational framework  describes the interaction between agents and their environment, with the aim of optimizing reward signals \citep{sutton2018}. The ultimate goal of RL is to discover the optimal policy, or the mapping between states and actions, that maximizes the expected cumulative reward over time. In this framework, agents learn to associate their actions with outcomes that lead to positive or negative rewards and use this association to optimize their policy. Beyond merely associating actions with rewards, advanced RL algorithms like AlphaZero demonstrate an unprecedented level of strategic depth and adaptability \citep{schrittwieser2020,silver2017}. These agents not only learn optimal policies linking states to actions for cumulative reward maximization, but also exhibit innovative strategies and problem-solving skills. Advanced RL algorithms can equip agents with optimal policies for navigating not only virtual environments like games but also the diverse and unpredictable settings of the real world. For instance, RL enhances robotics with advanced human-robot collaboration \citep{ghadirzadeh2021}, and aids autonomous vehicles in complex tasks like drone racing \citep{kaufmann2023}. Additionally, RL efficiently enables robots to learn complex tasks like walking in a park, showcasing its effectiveness in varied and dynamic environments \citep{smith2023}.

Although RL has made significant progress, especially in controlled environments, there are significant challenges in applying it to real-world scenarios \citep{dulac-arnold2021,thrun2014,zhang2020a, zhou2024}. The inherent structure of RL methods is challenged by issues such as sample efficiency, scalability, generalization, and robustness. One of the most significant challenges for RL in real-world applications is its sensitivity to noisy observations, as real-world sensors invariably contain some level of noise \citep{park2023,sun2023a,dulac-arnold2021}. Value-based approaches like Q-learning \citep{watkins1992} are  vulnerable to noisy observations \citep{fox2015a,moreno2006}. The issue of noise becomes more critical in the early stages of Q-learning, as it can significantly amplify errors and instability in Q-value estimates. This is particularly problematic during initial exploration phases, where the agent is still learning to form policy decisions based on these values. In the early stages of learning, small sample sizes can result in biased estimates, leading to incorrect decision-making. This can hinder the convergence of the learning process, necessitating a reset and relearning from a potentially better solution \citep{fox2015a}.

In contrast to the challenges faced by RL in real-world applications, animal intelligence exhibits remarkable efficiency and adaptability. Unlike RL algorithms, which can struggle with noisy data and require extensive training, animals demonstrate an innate ability to quickly adapt to changing environments. Both animals and humans require navigation and decision-making skills to locate resources, avoid dangers, and ultimately thrive in their environment. Mammals, birds, and reptiles use the hippocampus to encode spatial representations in order to organize and learn about relationships \citep{Rodrguez2002Spatial, Rodrguez2002Conservation,Fotowat2019Neural, Striedter2015Evolution}.

The discovery of place cells and grid cells in the hippocampus has led to the development of various hypotheses to explain how the brain processes spatial information \citep{okeefe1971, okeefe1978, hafting2005, fyhn2004, whittington2022}. One such theory is the predictive map theory proposed by Stachenfeld \citep{stachenfeld2017}, which is based on the successor representation (SR) learning algorithm. The concept of SR learning includes the dissection of rewards and state transitions \citep{dayan1993}. Consequently, SR learning involves representing the future expected occupancy of each state, under a particular policy. This representation enables the agent to efficiently compute the expected value of all states given a policy without carrying out an exhaustive search of the state space. 

Recent studies have highlighted the advantages of the successor features (SFs) approach, a linear approximation extension of the SR algorithm, since it decouples the feature representation from the reward function and is thus suitable for knowledge transfer between domains. In the paper that first coined the term SFs, they demonstrated that SFs facilitate the exchange of information across tasks and provide performance guarantees for the transferred policy even before learning begins \citep{barreto2017}. Another notable advancement is the universal successor features approximators, which combine the strengths of universal value function approximators, SFs, and generalized policy improvement \citep{borsa2018}. Furthermore, SFs bridges model-based and model-free RL by predicting future observation frequencies and facilitate generalization across different inputs \citep{lehnert2020}. Together, these studies demonstrate the usefulness and versatility of SFs in RL. 

Despite extensive research on transfer learning using SFs \citep{barreto2017,barreto2019, lehnert2020,borsa2018}, its robustness in noisy environments remains unexplored. While predecessor features (PFs) combining SFs and eligibility trace were proposed and shown to outperform SFs \citep{bailey2022, pitis2018}, their robustness advantage in noisy environments remains uninvestigated. Given that SR learning align with hippocampal activity patterns in animal intelligence, and recognizing that animals process state transitions and rewards separately during reward learning \citep{lee2012b,niv2009,geerts2020}, we hypothesize that SFs and PFs algorithms would exhibit greater robustness to noise compared to value-based learning approaches such as Q-learning or Q($\lambda$)-learning.


In this paper, we undertake a comparative analysis of the SFs and PFs algorithms against Q-learning and Q($\lambda)$-learning in environments with varying noise level. Section \ref{sec:background} provides a detailed explanation of the \emph{Markov decision processes} (MDPs) and delve into the mathematical and computational aspects of the SR learning and SFs algorithms. Following this, Section \ref{sec:pf} present the implementation of the PFs learning algorithm. Section \ref{sec:exp_study} elaborates on the establishment of noisy one-dimensional (1D) and two-dimensional (2D) grid world environments. This section analyzes and contrasts the performance of RL algorithms---the Q-learning, Q($\lambda$)-learning, SFs and PFs algorithms---in the noisy environments. Finally, Section \ref{sec:discussion} discusses the theoretical underpings that potentially contribute to the robustness of the SFs algorithm. This section also addresses the limitations of this study and elaborates the broader implications of our findings, particularly in the realms of artificial intelligence (AI) and neuroscience research.

\section{Background and Problem Formulation} \label{sec:background}

\subsection{Markov decision processes}
In this paper, we assume that RL agents interact with the environment through   \textit{Markov decision processes} (MDPs). The MDP used in this paper is a tuple $M := (\mathcal{S}, \mathcal{A}, R, \gamma)$ consisting of the following elements. The set $\mathcal{S}$ is the states (i.e., information about the space), and the set $\mathcal{A}$ is the space of actions that the agent can take. A function $R(s)$ maps the immediate reward received in state $s$. The discount factor $\gamma \in [0, 1)$ is a weight that reduces the value of future rewards. 

The main task of the RL agent is to find a policy function that maximizes the total discounted reward, also known as the return $G_{t} = \sum^{\infty}_{i=t}{\gamma^{i-t}R_{i+1}}$, where $R_{t} = R(S_{t})$. To solve this problem, we typically use \textit{dynamic programming} methods to define and compute a value function $V^{\pi}(s) := \mathbb{E}^{\pi}[G_{t} | S_{t} = s ]$ that depends on the policy $\pi$. The value function can be estimated by an approximation function $v^{\mathbf{w}}(s) \approx v^{\pi}(s)$ parameterized by a weight vector $\mathbf{w} \in \mathbb{R}^{d}$. To update the weight vector, TD learning can be utilized as follows: $\mathbf{w}_{t+1} = \mathbf{w}_{t} + \alpha[R_{t+1} + \gamma v_{\mathbf{w}}(s_{t+1}) - v_{\mathbf{w}}(s_{t})] \nabla_{\mathbf{w}} v_{\mathbf{w}}(s_{t})$.

TD(0) refers to an algorithm that uses the typical one-step TD update rule as mentioned above, while TD($\lambda$) refers to an classic algorithm that uses a eligibility trace based on past experience. The update rule for TD($\lambda$) is defined as $\mathbf{w}_{t+1} = \mathbf{w}_{t} + \alpha \delta_{t} \mathbf{e}_{t}$, where $\delta_{t} = R_{t+1} + \gamma v_{\mathbf{w}}(s_{t+1}) - v_{\mathbf{w}}(s_{t})$ is referred to TD error and $\mathbf{e}_{t} = \gamma \lambda \mathbf{e}_{t-1} + \nabla_{\mathbf{w}} v_{\mathbf{w}}(s_{t})$ is referred to the eligibility trace, where $\lambda$ indicates the trace decay parameter.

\subsection{Successor features learning}

The core idea of SR learning is that the value function can be decomposed into the expected visiting occupancy and reward of the successor state $s'$ as follows: $V^{\pi}(s) = \sum_{s'}\mathbb{E}^{\pi}[\sum_{i=t}^{\infty}\gamma^{i-t}\mathbb{I}(S_{i} = s')R(s') | S_{t} = s]=\sum_{s'} \mathbf{M}(s, s')R(s')$, where $\mathbb{I}(S_{i}=s')$ returns 1 if the agent visits the successor state $s'$ at time $t$, and 0 otherwise. Thus, $\mathbf{M}(s, :)$ represents the discounted expectation of a visit from state $s$ to it successor states, which can be called a successor state vector, or more generally, SFs.

Similar to how we used a weight vector $\mathbf{w}$ to estimate the value function, we can use a weight matrix and vector to estimate the $\mathbf{M}(s, :)$ and $R(s')$, respectively. In the tabular environment $\mathbb{R}^{|\mathcal{S}|}$, we can represent the state vector as a simple one-hot vector $\phi(s)$. Thus, we can factorize the reward vector as $R(s) = \phi(s) \cdot \mathbf{w}^{r}$ and the SFs as $\mathbf{M}(s, :) = \mathbf{W}^{sf}\phi(s) := \psi^{\mathbf{W}}(s)$. Therefore, we can rewrite the value function under the polity $\pi$ as follow: $V^{\pi}(s) = \psi^{\pi}(s) \cdot \bold{w}^{r}$, and the value approximation function as follow: $V^{\mathbf{W}}(s) = \psi^{\mathbf{W}}(s) \cdot \bold{w}^{r}$.

Using TD update rules, we can update $\mathbf{w}^{r}$ and $\mathbf{W}^{sr}$ as follows: 

\begin{equation}
\label{eq:r_vector}
\mathbf{w}_{t+1}^{r} = \mathbf{w}_{t}^{r} + \alpha_{r}( R_t - \phi(s) \cdot \mathbf{w}_{t}^{r})\phi(s)
\end{equation}

\begin{equation}
\label{eq:sr}
\mathbf{W}_{t+1}^{sf} = \mathbf{W}_{t}^{sf} + \alpha_{W}[\phi(s_{t}) + \gamma\psi^{\mathbf{W}}(s_{t+1}) - \psi^{\mathbf{W}}(s_{t})] \otimes \phi(s_{t})
\end{equation}

Therefore, we can present an example algorithms of SFs (Algorithm \ref{alg:sf}).

\begin{algorithm}
\caption{Successor feature learning}\label{alg:sf}
\begin{algorithmic}[1]
\State {\textbf{procedure}} \textsc{PF}($episodes, \mathbf{W}^{sf}, \mathbf{w}^{r}, \alpha_{W}, \alpha_{r}$)
\State \hspace{0.5cm} initialize $\mathbf{W}^{sf}, \mathbf{w}^{r}$
\State \hspace{0.5cm} \textbf{for} episode in 1...n \textbf{do}
\State \hspace{1cm} $s_{t} \gets $ initial state of episode
\State \hspace{1cm} \textbf{for} pair ($s_{t}, s_{t+1}$) and reward $r$ in episode \textbf{do}
\State \hspace{1.5cm} $\delta^{sf} \gets \phi(s_{t}) + \gamma\psi^{\mathbf{W}}(s_{t+1}) - \psi^{\mathbf{W}}(s_{t})$
\State \hspace{1.5cm} $\delta^{r} \gets r - \phi(s_{t+1}) \cdot \mathbf{w}^{r}$
\State \hspace{1.5cm} $\mathbf{W}^{sf} \gets \mathbf{W}^{sf} + \alpha_{W} \delta^{sf}$
\State \hspace{1.5cm} $\mathbf{w}^{r} \gets \mathbf{w}^{r} + \alpha_{r}\delta^{r} \phi(s_{t+1}) $
\State \hspace{0.5cm}\textbf{return}  $\mathbf{W}^{sf}, \mathbf{w}^{r}$
\end{algorithmic}
\end{algorithm}

\section{Predecessor Features Learning} \label{sec:pf}

Similarly to how the TD(0) algorithm can be converted into TD($\lambda$) by using the eligibility trace, the incorporation of SFs with the eligibility trace can result in the production of PFs. The eligibility trace can be updated by the one-hot state vector $\phi(s)$ as follow: $\mathbf{e}_{t} = \gamma\lambda\mathbf{e}_{t-1} + \phi(s_{t})$. Therefore, weight matrix of PFs update rule can be written as follows:
 
\begin{equation}
\label{eq:pr}
\mathbf{W}_{t+1}^{pf} = \mathbf{W}_{t}^{pf} + \alpha_{W}[\phi(s_{t}) + \gamma\psi^{\mathbf{W}}(s_{t+1}) - \psi^{\mathbf{W}}(s_{t})] \otimes \mathbf{e}_{t}
\end{equation}
where we used a linear approximation as $\psi^{\mathbf{W}}(s) = \mathbf{W}^{pf} \phi(s)$ for simplicity. It is worth noting that by using a one-hot state feature vector, we can consider PFs and PR learning to be highly analogous. Therefore, we can present an example algorithms of PFs (Algorithm \ref{alg:pf}).

\begin{algorithm}
\caption{Predecessor feature learning}\label{alg:pf}
\begin{algorithmic}[1]
\State {\textbf{procedure}} \textsc{PF}($episodes, \mathbf{W}^{pf}, \mathbf{w}^{r}, \lambda, \alpha_{W}, \alpha_{r}$)
\State \hspace{0.5cm} initialize $\mathbf{W}^{pf}, \mathbf{w}^{r}$
\State \hspace{0.5cm} \textbf{for} episode in 1...n \textbf{do}
\State \hspace{1cm} $s_{t} \gets $ initial state of episode
\State \hspace{1cm} $\mathbf{e} \gets 0$ (eligibility trace reset)
\State \hspace{1cm} \textbf{for} pair ($s_{t}, s_{t+1}$) and reward $r$ in episode \textbf{do}
\State \hspace{1.5cm} $\textbf{e} \gets \textbf{e} + \phi(s_{t})$
\State \hspace{1.5cm} $\delta^{pf} \gets \phi(s_{t}) + \gamma\psi^{\mathbf{W}}(s_{t+1}) - \psi^{\mathbf{W}}(s_{t})$
\State \hspace{1.5cm} $\delta^{r} \gets r - \phi(s_{t+1}) \cdot \mathbf{w}^{r}$
\State \hspace{1.5cm} $\mathbf{W}^{pf} \gets \mathbf{W}^{pf} + \alpha_{W} \delta^{pf} \otimes \mathbf{e}$
\State \hspace{1.5cm} $\mathbf{w}^{r} \gets \mathbf{w}^{r} + \alpha_{r}\delta^{r} \phi(s_{t+1}) $
\State \hspace{1.5cm} $\mathbf{e} \gets \gamma \lambda \mathbf{e}$
\State \hspace{0.5cm}\textbf{return}  $\mathbf{W}^{pf}, \mathbf{w}^{r}$
\end{algorithmic}
\end{algorithm}

Pitis \citep{pitis2018} introduced the concept of "source traces" as an application of eligibility traces to SR learning. They demonstrated the convergence of a TD($\lambda$)-like source learning algorithm and developed a novel algorithm for learning the source (SR) map, which outperformed previous approaches. Bailey \citep{bailey2022} also proposed an algorithm based on the same idea as proposed in this paper, naming it "predecessor features". The predecessor features algorithm was shown to be applicable to both tabular and feature representations, and demonstrated to outperform the "ExpectedTrace" \citep{hasselt2020} algorithm in the Cartpole task. However, it has not been compared to the SFs algorithm in a noisy environment.

\section{Experimental study} \label{sec:exp_study}
\subsection{Experimental design} \label{sec:method}
\subsubsection{Environments with noise}

\begin{figure}[H]
		\centering
    	\begin{tikzpicture}
    		\draw[draw=black, thick, solid] (-6.00,0.5) rectangle (-3.00,-0.5);
    		\node[black, anchor = north west] at (-5.2, 0.3) {\large Agent};
    		\draw[draw=black, thick, solid] (3, 0.5) rectangle (6, -0.5);
    		\node[black, anchor = north west] at (3.3, 0.3) {\large Environment};
    	    	\draw[draw=black, -latex, thin, dashed] (5.2, 0.5) -- (5.2, 1);
    	    	\node[black, anchor = south west] at (5.2, 0.5) {$s_{t+1}$};
    	    	\draw[draw=black, -latex, thin, dashed] (3.8, 0.5) -- (3.8, 1);
    	    	\node[black, anchor = south west] at (3.8, 0.5) {$r_{t+1}$};
    	    	\draw[draw=black, thin, dashed] (3,1) -- (6 ,1);
    	    	\draw[draw=black, thick, solid] (3.8, 1) -- (3.8, 1.5);
  	    	\node[black, anchor = south west] at (3.8, 1) {$r_{t}$};
    	    	\draw[draw=black, thick, solid] (-3.8, 1.5) -- (3.8, 1.5);
    	    	\draw[draw=black, -latex, thick, solid] (-3.8, 1.5) -- (-3.8, 0.5);
    	    	\node[black, anchor = south west] at (-3.8, 0.5) {$R_{t}$};
    	    	\draw[draw=black, thick, solid] (5.2, 1) -- (5.2, 2);
 	    	\node[black, anchor = south west] at (5.2, 1) {$s_{t}$};
 	    	\draw[draw=black, -latex, thick, solid] (5.2, 2) -- (0.2, 2);
 	    	\node[black, anchor = south west] at (2.2, 2) {$\phi(s_{t})$};
 	    	\node[black, anchor = south west] at (0.2, 2) {\footnotesize +};
 	    	\draw[draw=black, thick, solid] (0, 2) circle (0.2);
 	    	\draw[draw=black, -latex, thick, solid](0, 3) --(0, 2.2);
 	    	\node[black, anchor = south east] at (0, 2.2) {\footnotesize +};
 	    	\node[black, anchor = south east] at (0, 2.5) {noise $\boldsymbol{\epsilon}_t$};
 	    	\draw[draw=black, thick, solid] (-0.2, 2) -- (-5.2, 2);
 	    	\draw[draw=black, -latex, thick, solid] (-5.2, 2) -- (-5.2, 0.5);
 	    	\node[black, anchor = south west] at (-5.2, 0.5) {$\mathbf{o}_t$};
 	    	
			\draw[draw=black, thick, solid] (-4.5, -0.5) -- (-4.5, -1.0);
			\draw[draw=black, thick, solid] (-4.5, -1.0) -- (4.5, -1.0);
			\draw[draw=black, -latex, thick, solid] (4.5, -1.0) -- (4.5, -0.5); 
			\node[black, anchor = south] at (0, -1.0) {$a_{t}$};

	    \end{tikzpicture} 
       
    	\caption{Schematic representation of the environment-agent interaction in reinforcement learning with noisy observations. The agent receives an observation $\mathbf{o}_t$ and a reward $R_t$ at each time step $t$, based on which it decides on an action $a_t$. The environment, in turn, processes this action to update its state to $s_{t+1}$, providing the next state and reward ($r_{t+1}$) to the agent. A noise term $\boldsymbol{\epsilon}_t$ is added to the state representation $\phi(s_{t})$ before being fed back to the agent.}
	    \label{fig:NoisySchema}
    \end{figure}
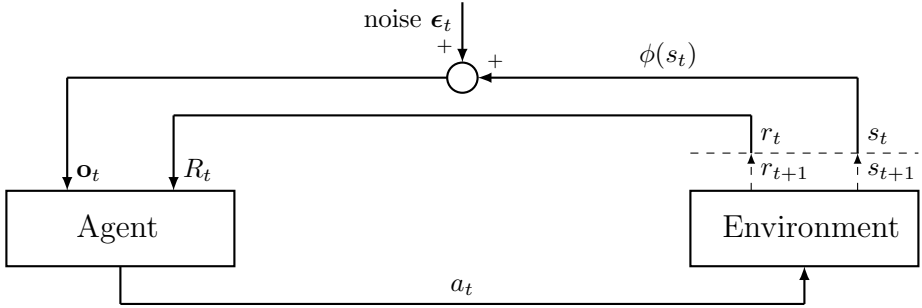

This section presents the noisy environment configuration adopted in our research. In our study, we consider a scenario within a MDP where the state is denoted by a one-hot vector $\phi(s_{t})$, indicative of the agents' positions within a grid world. We introduce a noisy MDP by adding a noise vector $\boldsymbol{\epsilon}_{t}$ to the $\phi(s_{t})$, thereby simulating the effect of uncertainty in the agents' perceived positions (Figure \ref{fig:NoisySchema}).

Following the approach recommended by \cite{zhang2020a} and \cite{sun2023a}, our study incorporated noise solely into the state observations, while maintaining unchanged state transition dynamics in the environment. This approach allows the noisy environment to more accurately model measurement errors, including sensor errors. To account for the effects of sensor noise commonly encountered in real-world scenarios, we added a Gaussian noise term to the observation vector $\mathbf{o}_{t}$ received by the agent in each state as following:

 \begin{equation}
 	\label{eq:noisy_obs}
 	\mathbf{o}_{t} = \phi(s_{t}) + \boldsymbol{\epsilon}_{t}, \hspace{0.2cm} \boldsymbol{\epsilon}_{t} \sim \mathcal{N}(\mathbf{0}, \sigma^2 \mathbf{I}) 
 \end{equation}
where $\boldsymbol{\epsilon}_{t}$ is the Gaussian noise vector with zero mean and covariance matrix $\sigma^2 \mathbf{I}$, where $\mathbf{I}$ is the identity matrix. Since the $\sigma$ regulates the level of noise in the $\mathbf{o}_{t}$, as $\sigma$ rises, the $\mathbf{o}_{t}$ grows noisier.

To evaluate the robustness of agents against the noisy observation $\mathbf{o}_{t}$, we conducted experiments in both a simple 1D grid world and a more complex 2D grid world. The observation vector's noise levels $\sigma$ were set at $[0.05, 0.25, 0.5]$ for low, medium, and high Gaussian noise to test the robustness of the agents. These levels were chosen to assess the algorithms' performance across a spectrum of environmental challenges, ensuring a comprehensive evaluation of their resilience to different degrees of stochastic disturbances. In our investigation, the discount factor $\gamma$ was set to 0.95.

	\begin{figure}[htpb]

		\centering
    	\begin{tikzpicture} 
	    
    		\node[anchor = north west] at (4, 2) {\includegraphics[width=5cm]{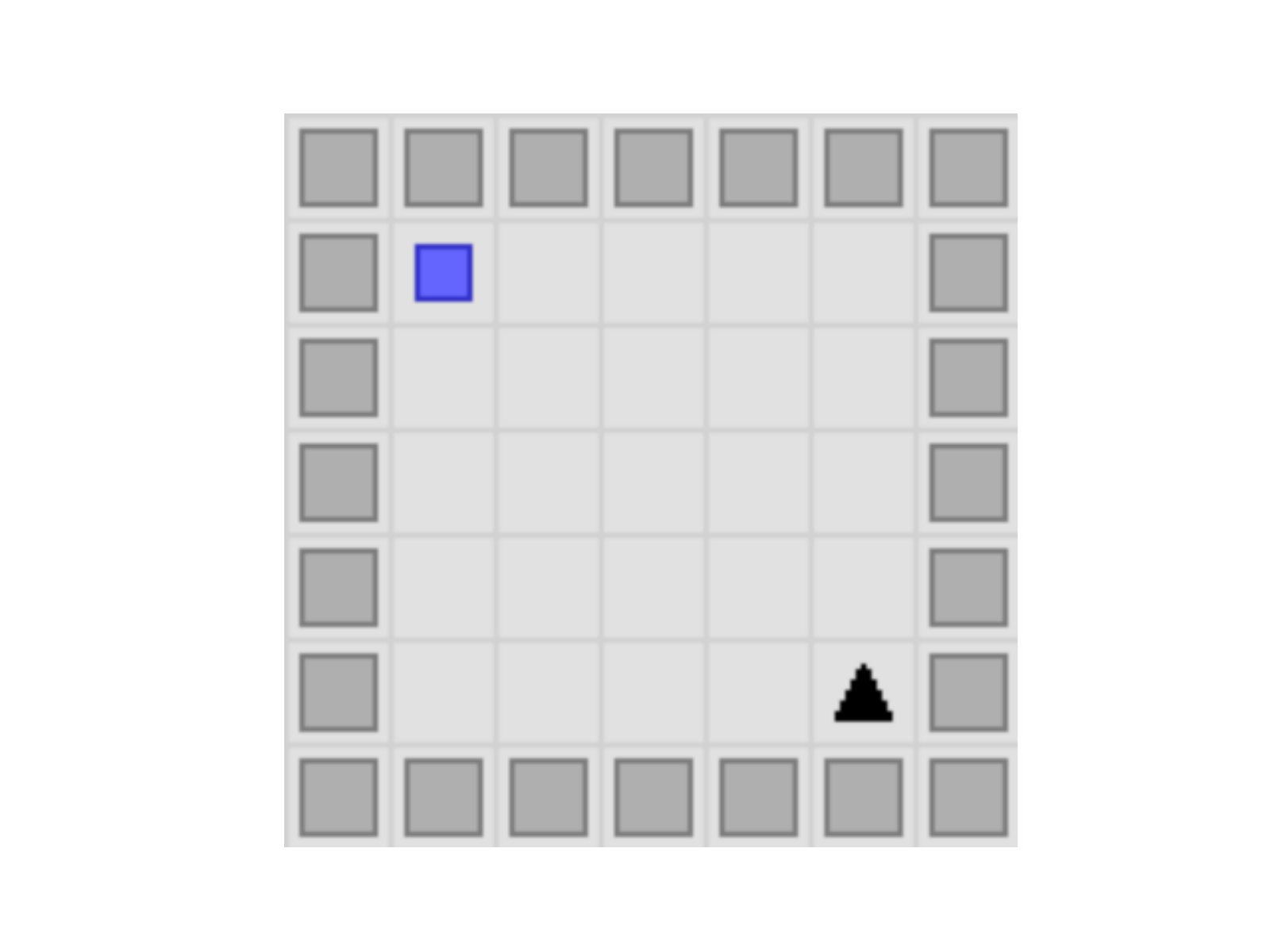}};
    		\node[black, anchor = south east] at (-0.5, 1) {A};
    		\draw node[rectangle, draw, minimum size=1cm] at (0,0) {1};
        	\draw node[rectangle, dotted, draw, minimum size=1cm] at (1,0) {......};
        	\draw node[rectangle, draw, minimum size=1cm] at (2,0) {$n$};
        	\draw node[rectangle, dotted, draw, minimum size=1cm] at (3,0) {......};
        	\draw node[rectangle, draw, minimum size=1cm] at (4,0) {$20$};
        	
        	\node[black, anchor = south east] at (5, 1) {B};

	    \end{tikzpicture}

    	\caption{Schematic of the grid world following the MDP. A. Illustrates a 1D grid world comprising 20 states, labeled from 1 to 20. The agent begins at state 1 (leftmost) and the goal is to reach state 20 (rightmost). B. A 2D grid world, structured as a 7x7 grid with barriers along the edges, restricting the agent's movement to a 5x5 area. The starting position is marked in the bottom right corner (triangle), and the goal is located in the top left corner (square).}
	    \label{fig:GridWorlds}
    \end{figure}
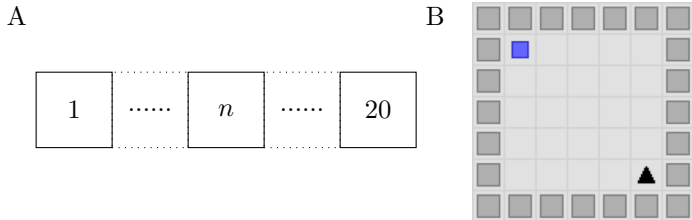

The 1D grid world in this study is modeled after the design used in our previous study \citep{lee2023d}. It consists of 20 states, where the agent starts at state 1 and aims to reach state 20 (Figure \ref{fig:GridWorlds}A). When the agent successfully reaches the target, it is awarded a reward of 1. If the target is not reached, the reward is 0. The agent's action space consists of two possible actions: move right or move left.

Our 2D grid world was developed using the neuro-nav library \citep{juliani2022a}, featuring a $7 \times 7$ grid with barriers along the edges, resulting in a $5\times5$ navigable space (Figure \ref{fig:GridWorlds}B). This design limits the agent's movable states to 25. However, the 2D grid presents a total of 49 observable states, adding complexity to the agent's task due to the increased number of potential states to observe and interpret. In this 2D grid world, the agent's objective is to navigate from the starting point in the bottom right corner to the goal located in the top left corner. Upon successfully reaching the target, the agent is awarded a reward of 1, otherwise 0. The agent's action space consists of four possible actions: move up, down, right or left.

\subsubsection{Algorithms hyperparameter}

In our methodology, we evaluated the performance of four RL algorithms---Q-learning, Q($\lambda$)-learning, SF, and PF---in a noisy environment. Q-learning and Q($\lambda$)-learning were selected due to benchmarking, providing a basis for comparison with SF and PF learning. For the Q($\lambda$)-learning and PF, we varied the $\lambda$ parameter at values of 0.7, 0.8, and 0.9 to determine its effect on the robustness of the algorithm under different levels of noise.

For the agent's policy function in our experiments, the $\epsilon$-greedy policy was implemented. This policy alternates between exploring random actions with a probability of $\epsilon$ and exploiting known actions with the highest Q-value estimate with a probability of $1-\epsilon$. To balance exploration and exploitation, the $\epsilon$ probability decays decayed over episodes according to $\epsilon_{k+1} := \text{max}(0.99 \cdot \epsilon_{k}, 0.01)$, where $\epsilon_{1} = 1.0$ and $k$ denoting the episode index.

For both Q-learning and Q($\lambda$)-learning, the learning rate was set to 0.1. Similarly, for SF and PF learning, the feature weight learning rate $\alpha_{W}$ and the reward vector learning rate $\alpha_{r}$ were both assigned to 0.1. 

\subsubsection{Evaluation metric}

To evaluate the performance of the agents in the noisy environments, we conducted a total of 100 runs of 3000 episodes each. In each episode, the agent starts from a starting state and continues working until it reaches a goal state, or up to 100 steps for a 1D grid world and up to 200 steps for a 2D grid world. We compared the performance of the agents by comparing the cumulative reward and the step length of the episode. We applied a moving average with a window of 20 episodes to the trajectory of episode length to minimize short-term fluctuations.

For the results of the 100 runs, we presented various statistical measures to encapsulate the stochastic nature inherent in the results. These include the mean, standard error of the mean (SEM), median, and the 25th and 75th percentiles. We also introduced the mean/SEM ratio as a critical metric. This approach provides a detailed analysis of the data, emphasizing central tendencies and variability, essential for identifying patterns and consistency in algorithm performance.

\subsection{Experimental Results}\label{sec:results}
\subsubsection{Comparison of RL Agents in a Noisy 1D Environment}

In this section, we present the results of our study comparing the performance of RL agents in a noisy 1D grid world. Our investigation into the robustness of various RL agents under different noise conditions reveals notable distinctions in performance, particularly for the SFs.

\begin{figure}[h]
\centering
\includegraphics[width=1\textwidth]{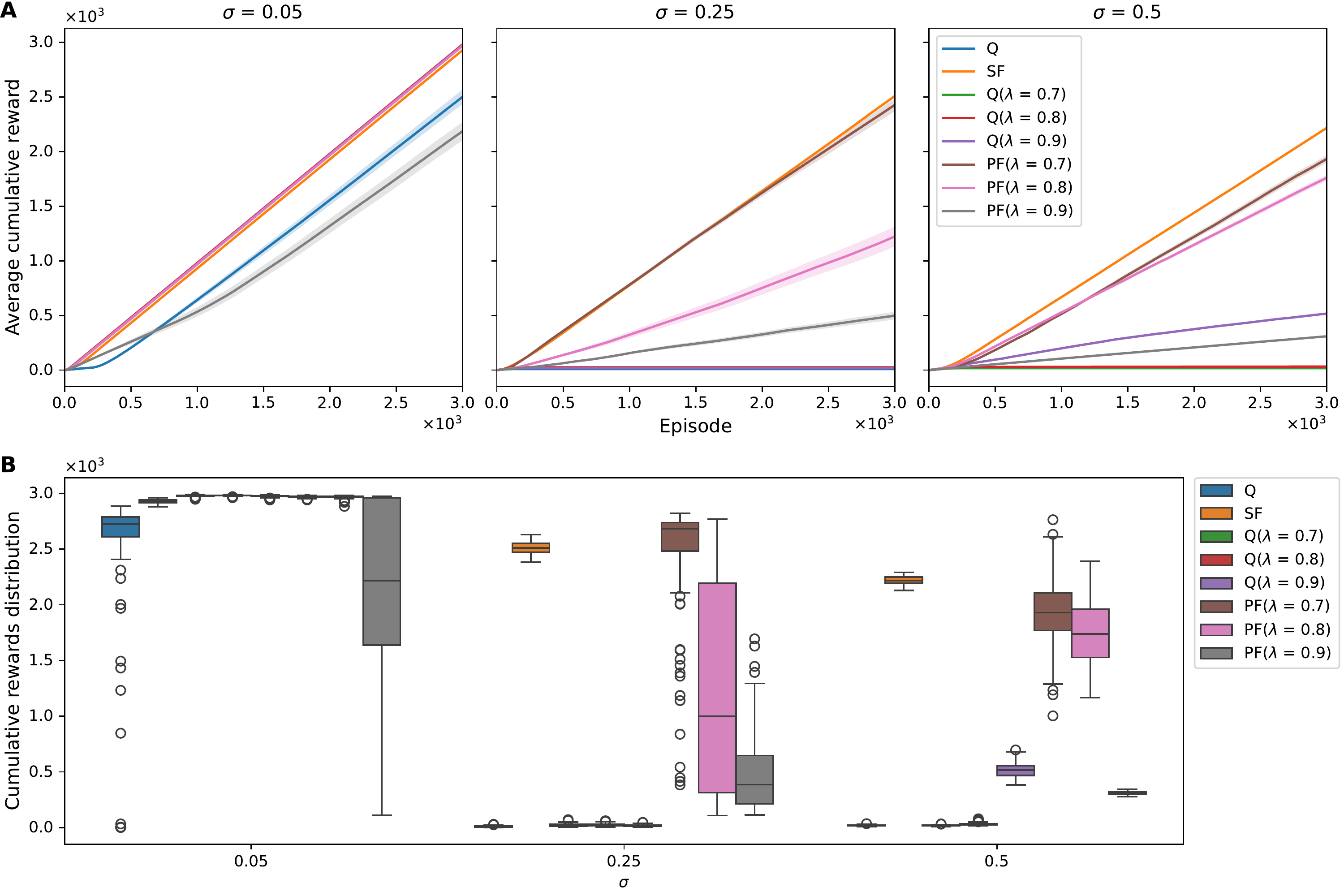}
\caption{Cumulative Reward of RL Agents in a Noisy 1D Environment. 
A. Average cumulative rewards over 3000 episodes for Q-learning, Q($\lambda$)-learning, SF, and PF under different levels of observation noise ($\sigma$ = 0.05, 0.25, 0.5). Each line represents the mean cumulative reward across episodes, with the shaded area depicting the standard error of the mean.
B. Distribution of cumulative rewards for each agent across noise settings. Box plots illustrate the median (central line), interquartile range (box limits), and outliers (individual points) for the cumulative rewards obtained over 3000 episodes, providing a comparative view of the reward distributions and the robustness of each algorithm to varying noise intensities.}
\label{fig:cumulative_reward}
\end{figure}

\paragraph{Robustness and Consistency of SF learning in Cumulative Reward Analysis}
In the 1D grid world with low Gaussian noise ($\sigma$ = 0.05), all agents exhibited relatively high mean cumulative rewards, with Q($\lambda$)-learning demonstrating the least variability as evidenced by the high mean/SEM ratios. However, the robustness of the SF agent becomes evident at increased noise levels ($\sigma$ = 0.25 and 0.50). Despite the environmental uncertainty introduced by higher noise, the SF agent maintained a stable performance curve (Figure \ref{fig:cumulative_reward}). This stability is further supported by the narrow interquartile ranges in Table \ref{tab:cumulative_reward}. At a noise level of $\sigma$ = 0.25, while PF with $\lambda$ = 0.7 exhibited a higher median cumulative reward than SF, the latter's tighter interquartile spread translated into a greater mean value and a higher mean/SEM ratio. In the 1D grid world with high Gaussian noise ($\sigma$ = 0.50), the SF algorithm not only preserved its performance, but it also degraded less than other algorithms. This robustness is particularly noteworthy in the mean/SEM ratio, which remained significantly higher for SF at $\sigma$ = 0.50 compared to Q-learning and PF algorithms. This metric emphasizes the robustness of SF in maintaining reward consistency under challenging noise conditions.

\begin{table}[h]
\centering
\begin{tabular}{|c|c|c|c|c|c|c|c|}
\hline
\textbf{Noise} & \textbf{RL agent} & \textbf{Mean} & \textbf{SEM} & \textbf{25\%} & \textbf{50\%} & \textbf{75\%} & \textbf{Mean/SEM} \\
\hline
0.05 & Q & 2500.85 & 66.05 & 2611.25 & 2724.0 & 2787.00 & 37.87  \\
& SF & 2928.07 & 1.80 & 2914.75 & 2931.0 & 2941.50 & 1625.43  \\
& Q($\lambda$ = 0.7) & 2978.54 & 0.66 & 2976.75 & 2980.0 & 2982.00 & 4479.62 \\
& Q($\lambda$ = 0.8) & \textbf{2980.05} & 0.41 & 2978.00 & \textbf{2981.0} & 2982.00 & \textbf{7289.12} \\
& Q($\lambda$ = 0.9) & 2974.02 & 0.82 & 2971.00 & 2975.0 & 2979.00 & 3614.60 \\
& PF($\lambda$ = 0.7) & 2968.45 & 0.83 & 2964.00 & 2970.0 & 2974.00 & 3579.95 \\
& PF($\lambda$ = 0.8) & 2968.10 & 1.37 & 2965.00 & 2971.0 & 2975.00 & 2173.90 \\
& PF($\lambda$ = 0.9) & 2186.20 & 83.63 & 1637.50 & 2217.0 & 2958.25 & 26.14 \\
\hline
0.25 & Q & 9.96 & 0.53 & 6.00 & 8.0 & 13.00 & 18.91 \\
& SF & \textbf{2508.13} & 5.79 & 2470.25 & 2510.5 & 2553.25 & \textbf{433.10} \\
& Q($\lambda$ = 0.7) & 22.62 & 1.23 & 13.75 & 21.0 & 28.50 & 18.35 \\
& Q($\lambda$ = 0.8) & 25.40 & 1.24 & 17.00 & 25.0 & 31.50 & 20.40 \\
& Q($\lambda$ = 0.9) & 18.39 & 0.88 & 12.00 & 18.0 & 24.00 & 21.01 \\
& PF($\lambda$ = 0.7) & 2426.04 & 58.05 & 2481.50 & \textbf{2681.5} & 2736.50 & 41.79 \\
& PF($\lambda$ = 0.8) & 1222.49 & 92.20 & 313.75 & 1000.0 & 2195.50 & 13.26 \\
& PF($\lambda$ = 0.9) & 498.26 & 35.63 & 214.75 & 384.5 & 647.50 & 13.98 \\
\hline
0.50 & Q & 19.22 & 0.57 & 15.00 & 18.0 & 22.00 & 33.71 \\
& SF & \textbf{2216.88} & 3.83 & 2193.75 & \textbf{2216.0} & 2248.25 & \textbf{578.62} \\
& Q($\lambda$ = 0.7) & 18.16 & 0.47 & 15.00 & 18.0 & 21.00 & 38.90 \\
& Q($\lambda$ = 0.8) & 32.11 & 1.09 & 25.00 & 30.5 & 35.00 & 29.38 \\
& Q($\lambda$ = 0.9) & 517.24 & 6.37 & 467.50 & 516.0 & 556.25 & 81.18 \\
& PF($\lambda$ = 0.7) & 1932.30 & 33.05 & 1768.25 & 1929.5 & 2108.00 & 58.47 \\
& PF($\lambda$ = 0.8) & 1760.41 & 27.90 & 1526.75 & 1738.5 & 1960.00 & 63.10 \\
& PF($\lambda$ = 0.9) & 309.23 & 1.54 & 298.00 & 308.0 & 321.00 & 201.04 \\
\hline
\end{tabular}
\caption{Assessment of Cumulative Reward Metrics Across Different Noise Levels in a Noisy 1D Environment for RL Algorithms. The table summarizes the cumulative rewards for various RL agents with noise levels $\sigma$ of 0.05, 0.25, and 0.5. The cumulative mean, the standard error of the mean (SEM), and quartiles (25\%, 50\%, 75\%) that map the distribution of cumulative rewards are shown. The Mean/SEM ratio provides insight into the consistency of reward relative to the average performance for each algorithm under the tested noise levels.}
\label{tab:cumulative_reward}
\end{table}

In contrast, the performance of Q-learning, Q($\lambda$)-learning, and PF algorithms, while relatively stable at the lower noise level, showed a pronounced decline as noise increased (Figure \ref{fig:cumulative_reward}). Even at the lowest noise level ($\sigma = 0.05$), Q-learning and PF with $\lambda = 0.9$ demonstrated suboptimal performance. This was evident from their lower mean cumulative rewards and higher SEM in outcomes compared to other algorithms, suggesting a lower level of robustness to even minimal environmental noise (Table \ref{tab:cumulative_reward}).

\begin{figure}[h]
\centering
\includegraphics[width=1\textwidth]{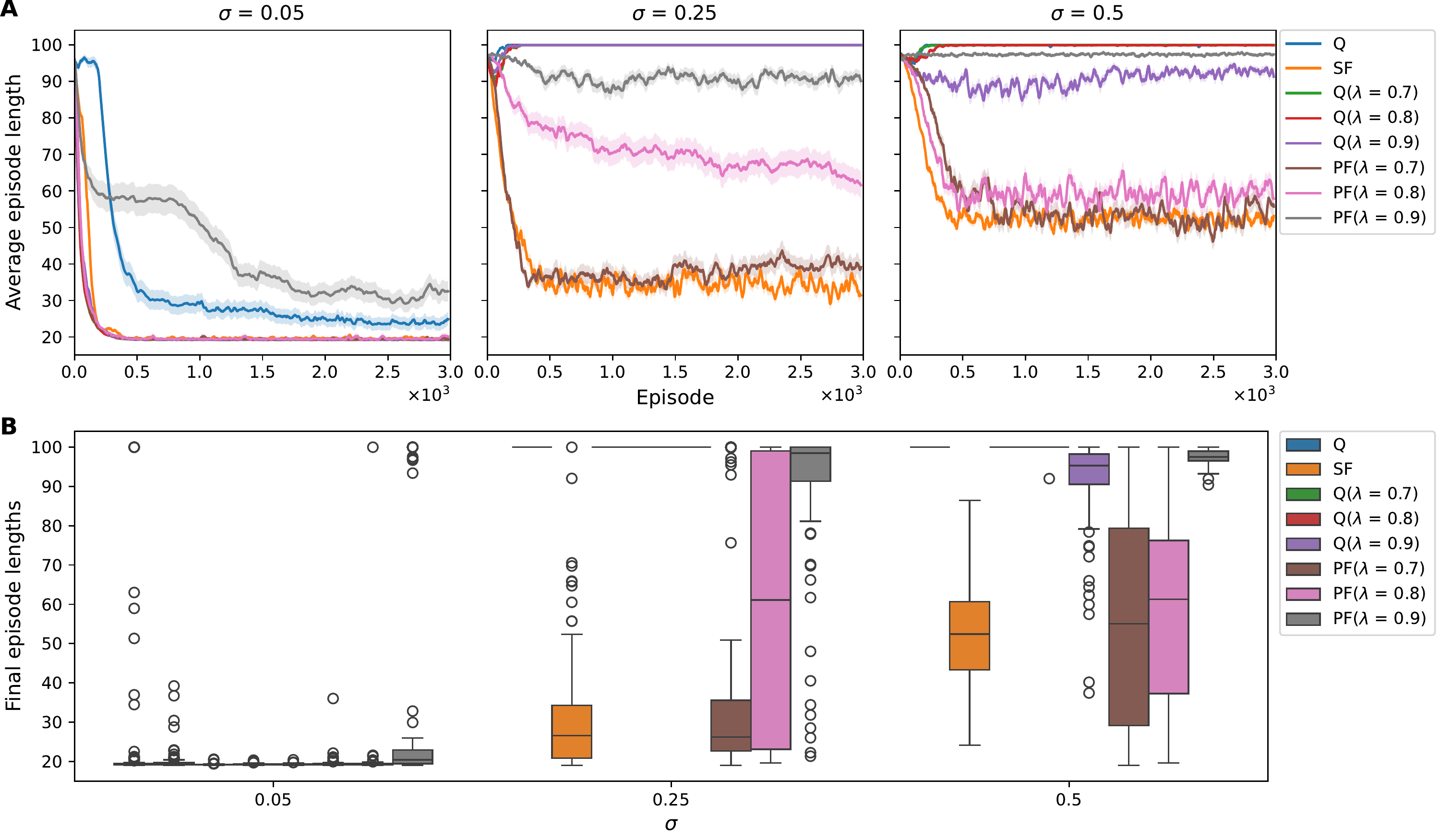}
\caption{Episode Length Trends and Distributions for RL Agents in a 1D Noisy Environment. A. Demonstrates the decreasing trend in average episode length across 3000 episodes for various algorithms, under noise levels $\sigma$ of 0.05, 0.25, and 0.5. The shaded areas represent the standard error of the mean. B. Demonstrates the distribution of moving-averaged episode lengths in the final episodes, capturing the stabilization of learning across various noise levels. Box plots illustrate the median (central line), interquartile range (box limits), and outliers (individual points).}
\label{fig:step_length}
\end{figure}

\paragraph{Stability of SF Learning in Reaching Goal in High Noise Environments}
Our investigation into algorithmic performance commenced with cumulative reward, subsequently followed by an examination of policy optimization efficiency as evidenced by the length of episode steps. In low-noise conditions, most algorithms rapidly approximated the optimal behavioral policy (the left panel of Figure \ref{fig:step_length}A). Q-learning and PF with $ \lambda = 0.9$, however, exhibited a slower convergence to the optimal path. With the introduction of increased noise, the disparity in performance became more pronounced. Q-learning and Q($\lambda$)-learning showed a notable decline in their ability to identify the optimal path consistently (Figure \ref{fig:step_length}).

In contrast, the SF and PF algorithms, particularly with $\lambda$ values of 0.7 and 0.8, demonstrated a robust ability to converge upon a shorter path even as noise intensified. Notably, SF maintained the smallest mean and variance in episode lengths across mid- and high-level noise environments (Table \ref{tab:episode_length}).

\begin{table}[h!]
\centering
\begin{tabular}{|c|c|c|c|c|c|c|}
\hline
\textbf{Noise} & \textbf{RL agent} & \textbf{Mean} & \textbf{SEM} & \textbf{25\%} & \textbf{50\%} & \textbf{75\%} \\
\hline
0.05 & Q & 24.87 & 1.87 & 19.10 & 19.23 & 19.42 \\
 & SF & 20.10 & 0.30 & 19.10 & \textbf{19.20} & 19.65 \\
 & Q($\lambda$ = 0.7) & \textbf{19.20} & 0.02 & 19.10 & \textbf{19.20} & 19.21 \\
 & Q($\lambda$ = 0.8) & 19.22 & 0.02 & 19.10 & \textbf{19.20} & 19.30 \\
 & Q($\lambda$ = 0.9) & 19.22 & 0.02 & 19.10 & \textbf{19.20} & 19.30 \\
 & PF($\lambda$ = 0.7) & 19.52 & 0.17 & 19.10 & \textbf{19.20} & 19.40 \\
 & PF($\lambda$ = 0.8) & 20.15 & 0.81 & 19.10 & 19.23 & 19.41 \\
 & PF($\lambda$ = 0.9) & 32.52 & 2.81 & 19.40 & 20.40 & 22.91 \\
\hline
0.25 &Q & 100.00 & 0.00 & 100.00 & 100.00 & 100.00 \\
 & SF & \textbf{31.38} & 1.57 & 20.79 & 26.52 & 34.27 \\
 & Q($\lambda$ = 0.7) & 100.00 & 0.00 & 100.00 & 100.00 & 100.00 \\
 & Q($\lambda$ = 0.8) & 100.00 & 0.00 & 100.00 & 100.00 & 100.00 \\
 & Q($\lambda$ = 0.9) & 100.00 & 0.00 & 100.00 & 100.00 & 100.00 \\
 & PF($\lambda$ = 0.7) & 39.14 & 2.81 & 22.68 & \textbf{26.17} & 35.54 \\
 & PF($\lambda$ = 0.8) & 61.60 & 3.63 & 23.10 & 61.10 & 99.08 \\
 & PF($\lambda$ = 0.9) & 90.13 & 1.92 & 91.31 & 98.50 & 100.00 \\
\hline
0.50 & Q & 100.00 & 0.00 & 100.00 & 100.00 & 100.00 \\
 & SF & \textbf{53.13} & 1.28 & 43.29 & \textbf{52.42} & 60.64 \\
 & Q($\lambda$ = 0.7) & 100.00 & 0.00 & 100.00 & 100.00 & 100.00 \\
 & Q($\lambda$ = 0.8) & 99.92 & 0.08 & 100.00 & 100.00 & 100.00 \\
 & Q($\lambda$ = 0.9) & 91.30 & 1.19 & 90.54 & 95.30 & 98.28 \\
 & PF($\lambda$ = 0.7) & 55.73 & 2.68 & 29.15 & 55.02 & 79.39 \\
 & PF($\lambda$ = 0.8) & 57.90 & 2.36 & 37.25 & 61.25 & 76.25 \\
 & PF($\lambda$ = 0.9) & 97.41 & 0.20 & 96.54 & 97.53 & 99.00 \\
\hline
\end{tabular}
\caption{Distribution of Moving-averaged Episode Lengths in the Final Episodes Across Different Noise Levels in a Noisy 1D Environment. The table summarizes the episode length distributions for various RL agents in the final episodes, under noise levels $\sigma$ of 0.05, 0.25, and 0.50. The mean episode lengths, the standard error of the mean (SEM), and the quantiles (25\%, 50\%, 75\%) for each algorithm are shown.}
\label{tab:episode_length}
\end{table}

\begin{figure}[h]
\centering
\includegraphics[width=1\textwidth]{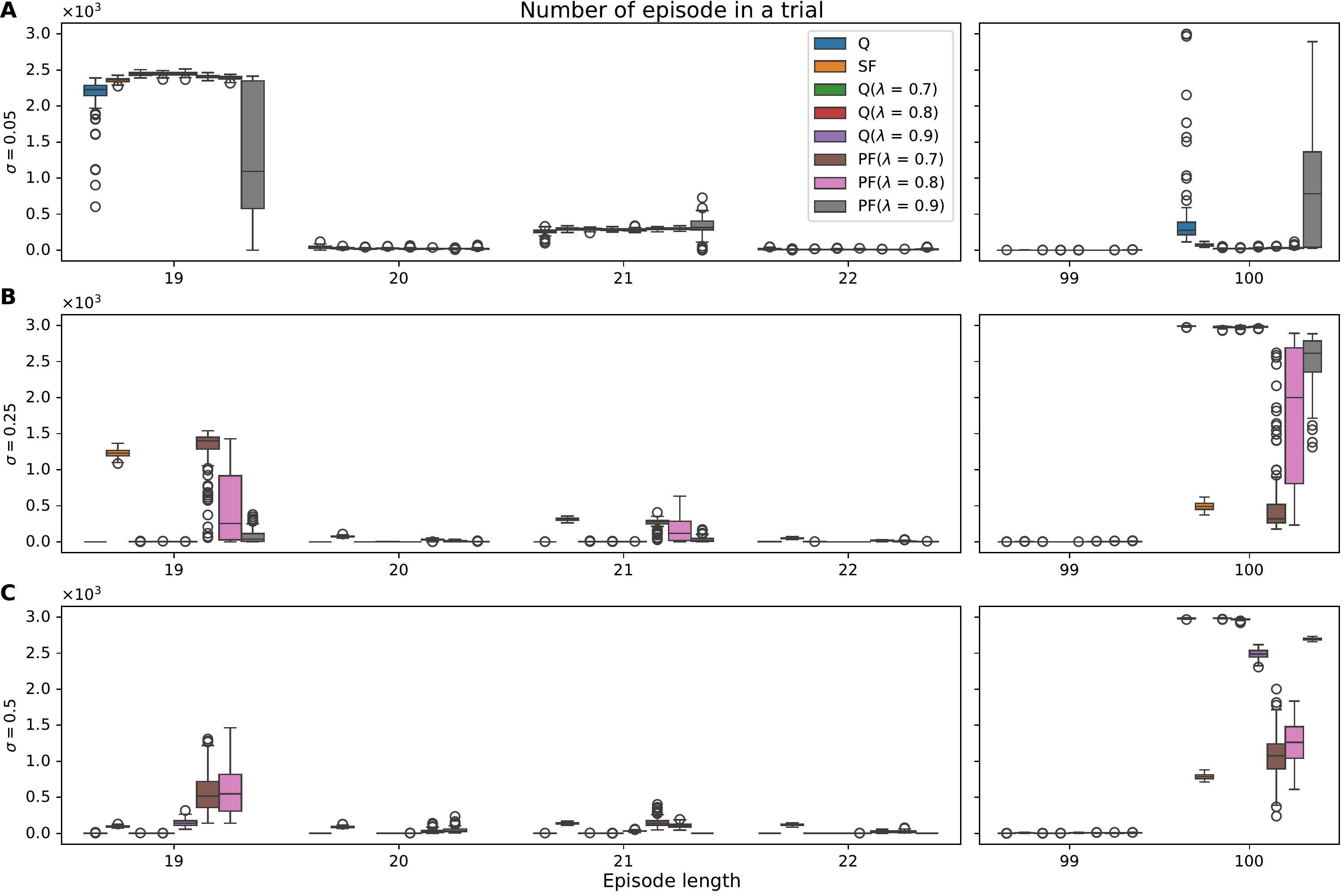}
\caption{Comparison of RL Algorithms for a Specific Range of Episode Lengths in a Noisy 1D Environment. This figure illustrates the frequency distribution of episode lengths across 3000 trials for various RL algorithms, categorized by noise levels with  values of 0.05 (A), 0.25 (B), and 0.5 (C). The visualization concentrates on episodes ending with fewer than 22 steps or exceeding 99 steps. Box plot represent the distribution of the episode count for a given length, providing a visual comparison of algorithm efficiency and consistency.}
\label{fig:step_dist}
\end{figure}

We focused on the distribution of episodes within a trial that either rapidly converged to an optimal step length or failed to do so, culminating in the maximum allowed step length of 100 (Figure \ref{fig:step_dist}). Our initial observation across the entire span of episode lengths indicated a rarity of episodes falling within the mid-range --- specifically, those in the lower 30s to the high 90s were infrequent (Figure \ref{fig:whole_step_dist}). Consequently, our analysis was refined to emphasize the extremes, presenting a contrast between episodes that efficiently reached the goal and those that did not.

In low noise environments ($\sigma$ = 0.05), a pronounced proportion of episodes is observed at lower step lengths for all algorithms, indicating a rapid convergence to the optimal policy. However, a discernible spread towards the 100-step length for Q-learning and PF with $\lambda$ = 0.9 reflects a less efficient navigation towards the goal (the right panel of Figure \ref{fig:step_dist}A).

As noise levels rise, the distribution of episode lengths for Q-learning and Q($\lambda$) becomes more scattered, with an increased frequency of episodes reaching the 100-step length. In contrast, the SF and the PF with $\lambda$ =  0.7 and 0.8, stood out by consistently achieving a higher frequency of episodes with shorter step lengths within medium ($\sigma$ = 0.25) and high ($\sigma$ = 0.5) noise levels. In the high ($\sigma$ = 0.5) noise condition, while the SF algorithm exhibited a reduced incidence of achieving the optimal step length, it notably also displayed fewer episodes reaching the upper bound of 100 steps (the right panel of Figure \ref{fig:step_dist}C). This pattern suggests that, despite the challenging conditions, the SF algorithm reliably navigates towards the goal within the step constraints imposed by the environment.

\begin{figure}[h]
\centering
\includegraphics[width=1\textwidth]{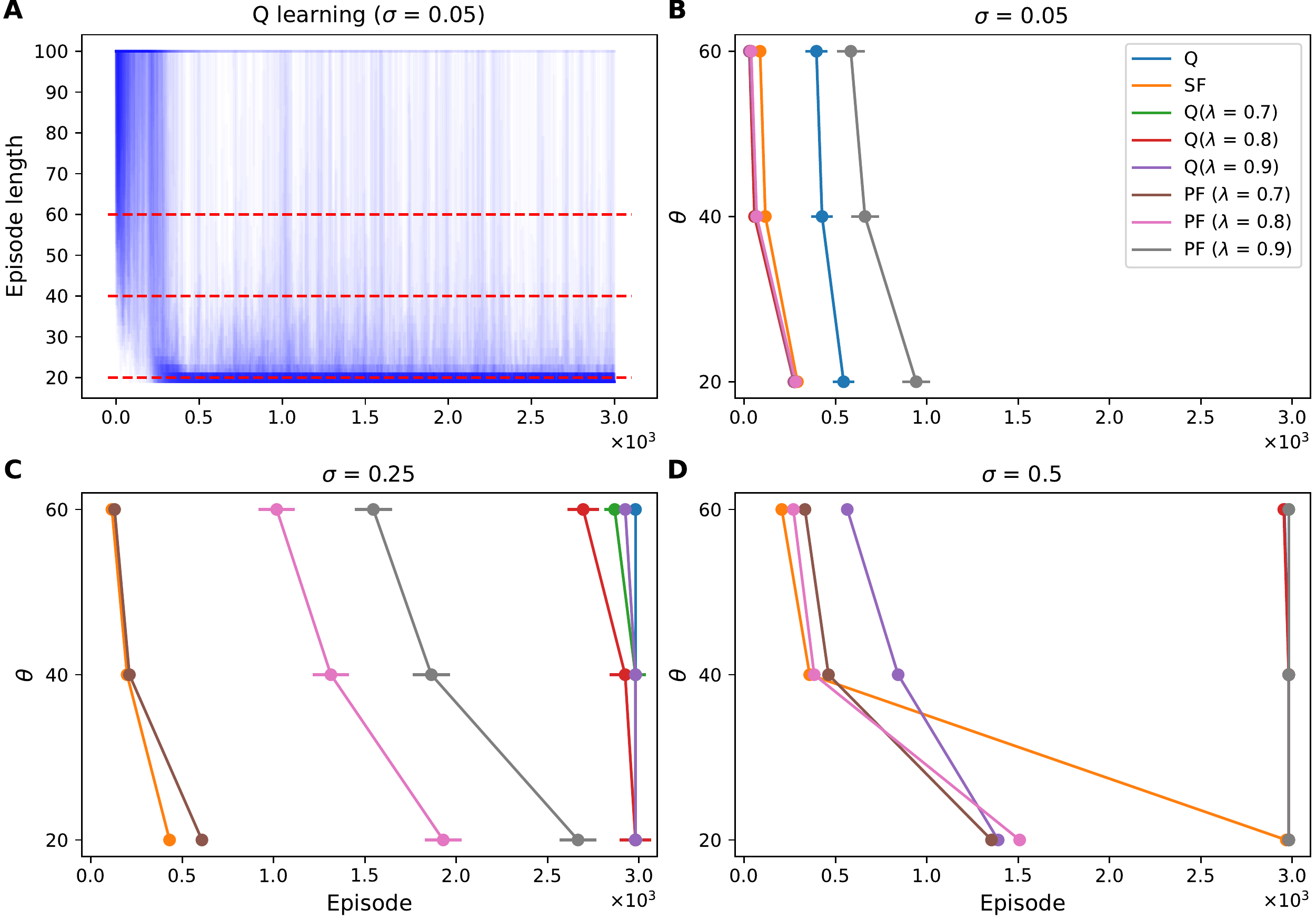}
\caption{Convergence Trends of RL Algorithms Under Variable Noise Condition. A. Visualization of 100 episode length trajectories for Q-learning at a low noise level ($\sigma$ = 0.05), with each trajectory represented by a line with an alpha transparency of 0.01. This allows for the visualization of the density of episode lengths over 3000 episodes, with overlapping trajectories appearing darker. B-D:  These line plots chart the convergence trajectory of the RL agents, displaying the average episode number at which each algorithm first attained a predefined performance threshold ($\theta$) across varying noise intensities: low (B), medium (C), and high (D). The markers denote the mean episode number where the respective algorithm's performance first met the threshold, offering a measure of its learning speed and adaptability to noise. The error bars represent the standard error of the mean.}
\label{fig:step_threshold_line}
\end{figure}

\paragraph{Policy Convergence Efficacy Under Varying Noise Conditions: A Threshold-Based Approach}
To assess how RL agents achieve optimal convergence in the initial episodes within a 1D grid world composed of 20 states, we compared the first episode that met a specific performance threshold ($\theta$) across each trial. Ideally, we would establish a performance threshold for the RL agents to reach the goal state within 20 steps or fewer. However, as environmental noise increased, the frequency of agents reaching this optimal number of steps decreased (Figure \ref{fig:step_dist}). This necessitated the extension of our analysis to include less stringent thresholds, accounting for the first episodes completed within 40 and 60 steps or fewer (Figure \ref{fig:step_threshold_line}).

In the low noise ($\sigma$ = 0.05), the data shows that most algorithms, excluding Q-learning and PF with $\lambda$ = 0.9, efficiently converge to an episode length of fewer than 20 steps within the initial few hundred episodes (Figure \ref{fig:step_threshold_line}B). However, we can observe a divergence in algorithmic performance, as we increase the environmental noise to $\sigma$ = 0.25. SF learning and PF with $\lambda$ = 0.7 stand out by reaching the optimal policy in early episodes, demonstrating their robustness. In contrast, Q-learning and Q($\lambda$)-learning do not exhibit this efficiency, failing to converge to the optimal policy within the 3000-episode timeframe (Figure \ref{fig:step_threshold_line}C).

At the highest noise level ($\sigma$ = 0.5), PF with $\lambda$ = 0.7 and 0.8, as well as Q($\lambda$)-learning with $\lambda$ = 0.9, maintain a commendable learning rate. SF learning, while not achieving the 20-step threshold, does reach the 40 and 60 steps thresholds more rapidly than other algorithms (Figure \ref{fig:step_threshold_line}D). Taken together, these results suggest that the learning efficiency and robustness of the RL algorithm to noise is non-linear. Nevertheless, SF learning showed efficiency and robustness in reaching less stringent step thresholds quickly at all noise levels.

\subsubsection{Comparison of RL Agents in a Noisy 2D Environment}

Our investigation extended into the realm of higher-dimensional noise environments to ascertain if the performance trends observed in 1D grid worlds would persist. We engaged in a series of experiments within a 2D grid world, where the agent navigates a 7x7 observable space, with the actual movable area being a 5x5 grid due to peripheral walls (Figure \ref{fig:GridWorlds}B). The action space was expanded to four possible moves--up, down, left, and right. Despite the optimal path length being a mere eight steps, we adjusted the episode length limit to 200 to account for the increased complexity arising from a larger action space and observable world size.

\begin{figure}[h]
\centering
\includegraphics[width=1\textwidth]{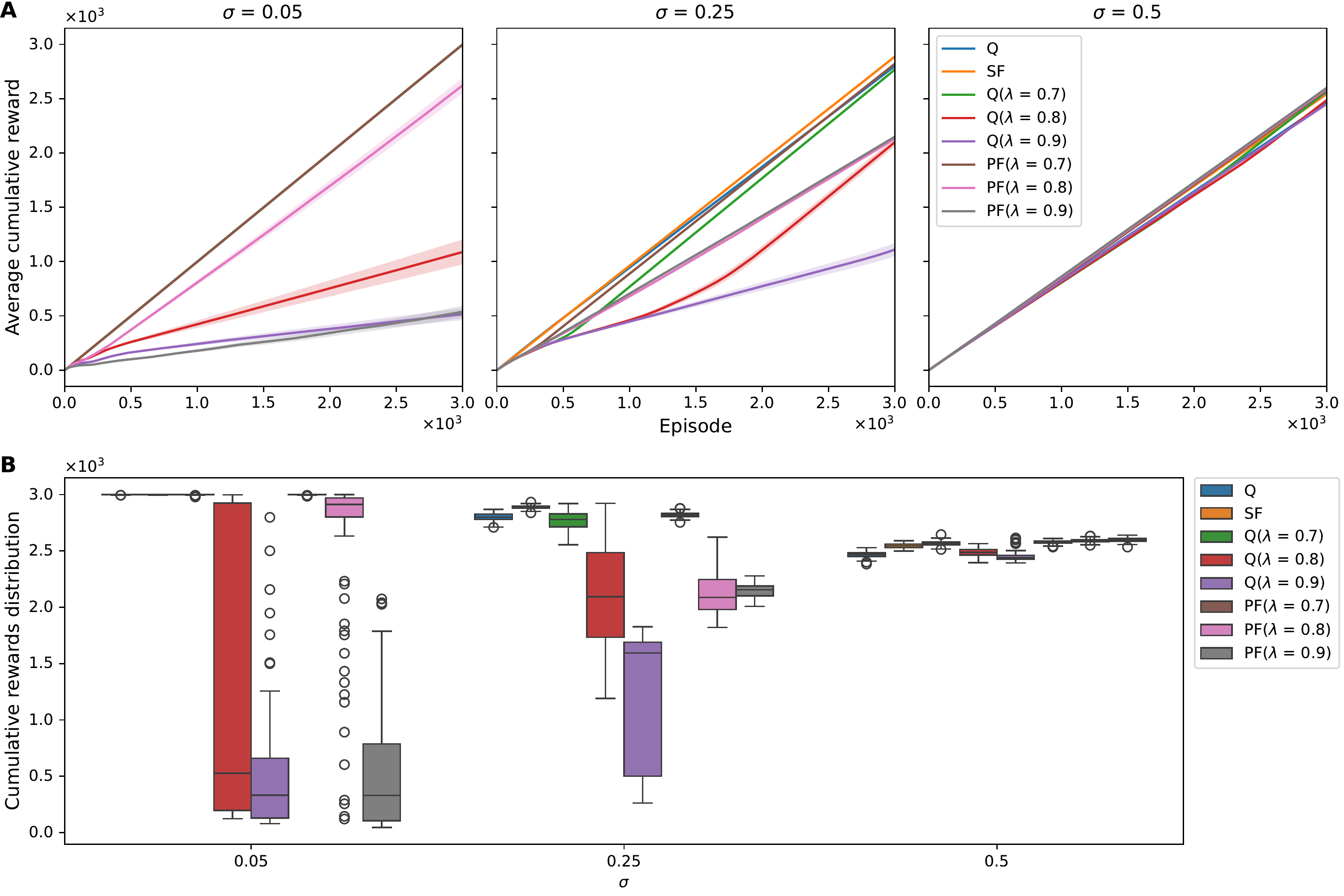}
\caption{Cumulative Reward of RL Agents in a Noisy 2D Environment. 
A. Average cumulative rewards over 3000 episodes for Q-learning, Q($\lambda$)-learning, SF, and PF under different levels of observation noise ($\sigma$ = 0.05, 0.25, 0.5). Each line represents the mean cumulative reward across episodes, with the shaded area depicting the standard error of the mean.
B. Distribution of cumulative rewards for each agent across noise settings. Box plots illustrate the median (central line), interquartile range (box limits), and outliers (individual points) for the cumulative rewards obtained over 3000 episodes, providing a comparative view of the reward distributions and the robustness of each algorithm to varying noise intensities.}
\label{fig:2D_cumulative_reward}
\end{figure}

\paragraph{Anomalous Performance Enhancements in Cumulative Reward Acquisition by RL Agents}
We conducted a comparative analysis of the cumulative rewards achieved by different RL agents at noise levels $\sigma$ of 0.05, 0.25, and 0.50, as depicted in Figure \ref{fig:2D_cumulative_reward} and Table \ref{tab:2D_cumulative_reward}. At the low noise level ($\sigma$ = 0.05), the Q-learning, Q($\lambda$ = 0.7), SF, and PF($\lambda$ = 0.7) agents demonstrate near-optimal performance hovering close to the maximum possible reward, as evidenced by the negligible SEM. The PF learning with $\lambda$ value of 0.8 shows poor performance with relatively low average reward and high SEM. In contrast, at this noise level, Q($\lambda$ = 0.8, 0.9)-learning and PF($\lambda$ = 0.9) learning  did not perform as well, exhibiting a lower mean reward and elevated SEM, indicative of lesser stability in reward acquisition. As the noise level increases to $\sigma$ = 0.25, a surprising uptick in accumulated rewards was noted for these latter algorithms, suggesting a possible noise threshold that might facilitate enhanced exploratory behaviors or other noise-dependent dynamics not yet fully understood. SF learning, however, consistently maintained superior performance, with a relatively Mean/SEM ratio, suggesting a resilient performance despite the heightened noise.

At high noise levels ($\sigma$ = 0.5), an unexpected pattern emerged where all algorithms seemed to converge towards the upper echelons of performance, a phenomenon warranting further investigation.

\begin{table}[h]
\centering
\begin{tabular}{|c|c|c|c|c|c|c|c|}
\hline
\textbf{Noise} & \textbf{RL agent} & \textbf{Mean} & \textbf{SEM} & \textbf{25\%} & \textbf{50\%} & \textbf{75\%} & \textbf{Mean/SEM} \\
\hline
0.05 & Q & 2998.76 & 0.13 & 2998.00 & \textbf{2999.00} & 3000.00 & 23870.59 \\
 & SF & \textbf{2998.88} & 0.10 & 2998.00 & \textbf{2999.00} & 3000.00 & \textbf{28637.94} \\
 & Q($\lambda$ = 0.7) & 2997.18 & 0.37 & 2997.00 & 2998.00 & 2999.00 & 8008.11 \\
 & Q($\lambda$ = 0.8) & 1087.93 & 115.40 & 196.75 & 526.50 & 2924.25 & 9.43 \\
 & Q($\lambda$ = 0.9) & 514.82 & 54.41 & 128.50 & 332.50 & 660.50 & 9.46 \\
 & PF($\lambda$ = 0.7) & 2997.40 & 0.30 & 2997.00 & 2998.00 & 2999.00 & 10088.04 \\
 & PF($\lambda$ = 0.8) & 2622.01 & 70.16 & 2800.25 & 2912.50 & 2970.25 & 37.37 \\
 & PF($\lambda$ = 0.9) & 537.09 & 54.58 & 104.50 & 330.00 & 786.75 & 9.84 \\
\hline
0.25 & Q & 2798.16 & 3.54 & 2779.75 & 2798.50 & 2826.50 & 789.50 \\
 & SF & \textbf{2886.03} & 1.63 & 2877.75 & \textbf{2886.00} & 2896.25 & \textbf{1770.18} \\
 & Q($\lambda$ = 0.7) & 2766.18 & 8.09 & 2713.25 & 2779.50 & 2828.25 & 342.05 \\
 & Q($\lambda$ = 0.8) & 2097.67 & 41.30 & 1734.25 & 2094.00 & 2485.00 & 50.79 \\
 & Q($\lambda$ = 0.9) & 1109.58 & 61.72 & 500.50 & 1595.00 & 1689.75 & 17.98 \\
 & PF($\lambda$ = 0.7) & 2819.26 & 2.44 & 2805.00 & 2819.50 & 2832.25 & 1154.06 \\
 & PF($\lambda$ = 0.8) & 2133.44 & 20.30 & 1979.75 & 2087.50 & 2246.75 & 105.08 \\
 & PF($\lambda$ = 0.9) & 2148.33 & 6.01 & 2102.00 & 2154.00 & 2189.25 & 357.27 \\ \hline
0.50 & Q & 2466.29 & 2.54 & 2451.00 & 2469.00 & 2482.00 & 971.20 \\
 & SF & 2543.89 & 2.06 & 2528.75 & 2544.50 & 2560.00 & 1232.87 \\
 & Q($\lambda$ = 0.7) & 2565.21 & 2.28 & 2551.75 & 2565.50 & 2577.50 & 1126.23 \\
 & Q($\lambda$ = 0.8) & 2486.37 & 3.36 & 2462.75 & 2487.00 & 2513.25 & 739.47 \\
 & Q($\lambda$ = 0.9) & 2451.33 & 4.73 & 2426.00 & 2440.00 & 2460.00 & 518.03 \\
 & PF($\lambda$ = 0.7) & 2577.74 & 1.66 & 2569.25 & 2580.00 & 2588.00 & \textbf{1554.31} \\
 & PF($\lambda$ = 0.8) & 2590.20 & 1.70 & 2579.75 & 2590.00 & 2599.25 & 1524.63 \\
 & PF($\lambda$ = 0.9) & \textbf{2599.42} & 1.92 & 2585.75 & \textbf{2601.50} & 2613.00 & 1351.66 \\
 \hline
\end{tabular}
\caption{Assessment of Cumulative Reward Metrics Across Different Noise Levels in a Noisy 2D Environment for RL Algorithms. The table summarizes the cumulative rewards for various RL agents with noise levels $\sigma$ of 0.05, 0.25, and 0.5. The cumulative mean, the standard error of the mean (SEM), and quartiles (25\%, 50\%, 75\%) that map the distribution of cumulative rewards are shown. The Mean/SEM ratio provides insight into the consistency of reward relative to the average performance for each algorithm under the tested noise levels.}
\label{tab:2D_cumulative_reward}
\end{table}

\paragraph{Anomalous Episode Length Trends in 2D RL}
Our examination into the 2D grid world's noise impact on RL algorithms extended to an in-depth analysis of episode lengths to clarify some unexpected trends observed in cumulative reward outcomes. 

\begin{figure}[h]
\centering
\includegraphics[width=1\textwidth]{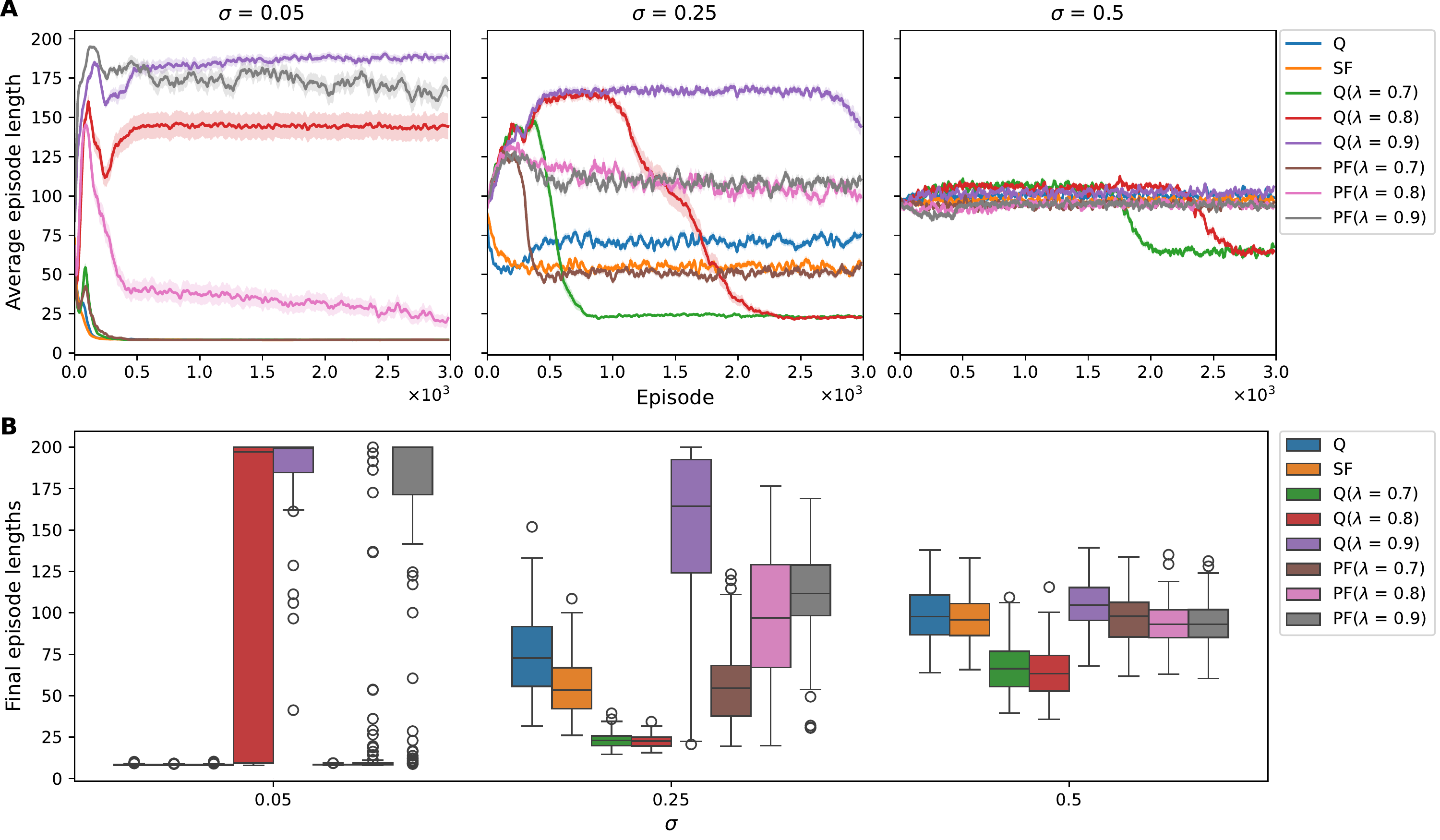}
\caption{Episode Length Trends and Distributions for RL Agents in a 2D Noisy Environment. A. Demonstrates the decreasing trend in average episode length across 3000 episodes for various algorithms, under noise levels $\sigma$ of 0.05, 0.25, and 0.5. The shaded areas represent the standard error of the mean. B. Demonstrates the distribution of moving-averaged episode lengths in the final episodes, capturing the stabilization of learning across various noise levels. Box plots illustrate the median (central line), interquartile range (box limits), and outliers (individual points).}
\label{fig:2D_step_length}
\end{figure}

In Figure \ref{fig:2D_step_length}A, we illustrate the trajectory of the average episode lengths over the span of 3000 episodes within a 2D noisy environment. In addition, in Figure \ref{fig:2D_step_length}B, the box plot elucidates the distribution of moving-averaged episode lengths in the final episodes, revealing the degree of convergence stability exhibited by the agents. Table \ref{tab:2D_episode_length} complements this visual representation by offering a statistical breakdown of these episode lengths. 

At a low noise level ($\sigma = 0.05$), algorithms such as Q-learning, SF, Q($\lambda = 0.7$), and PF($\lambda = 0.7$) exhibit rapid alignment with the optimal path, indicating fast policy adaptation. The distribution of moving-averaged episode lengths in the final episodes reveals that the algorithms efficiently approximate the environment's optimal path length. The significantly low SEM values underscore the algorithms' consistent precision, highlighting their stable performance with minimal variability.

Conversely, algorithms like PF($\lambda = 0.8$) initially manifest an increase in episode lengths, hinting at initial policy inefficiencies. Nonetheless, this is mitigated over time by a decelerating trend in episode length, denoting incremental policy refinement. In the final episode, the distribution of moving-averaged episode lengths indicates that the majority, specifically the top 75\%, align closely with the optimal path length, although there are notable outliers.

In contrast, however, Q($\lambda$ = 0.8, 0.9)-learning and PF($\lambda$ = 0.9) learning displayed a less adaptive response, with their learning curves plateauing or even worsening, suggesting a difficulty in overcoming the initial suboptimal policies. 

As the noise level increased to $\sigma$ = 0.25, a distinct pattern emerged for the Q-learning and SF algorithms: an initial reduction in episode lengths suggested promising policy adaptation, yet this progress plateaued, failing to reach optimal levels. Algorithms employing eligibility traces diverged; they first exhibited an upsurge in average episode lengths, indicative of initial policy inefficiencies, followed by a gradual reduction. Notably, Q($\lambda$ = 0.7, 0.8) algorithms achieved comparatively shorter episode lengths, and Q($\lambda$ = 0.8) outperformed its lower noise condition result. This trend reflective of their differential response to increased environmental stochasticity.

At high noise levels examined ($\sigma$ = 0.5), there is a general tendency for all algorithms to not significantly deteriorate or improve and to remain unchanged around a step length of 100 at the start of training. Nonetheless, Q($\lambda$)-learning, particularly at $\lambda$ values of 0.7 and 0.8, demonstrates a reduction in average episode length during the latter stages of learning. This observation hints at a potential robustness to noise or an adaptive exploration strategy.

\begin{table}[h!]
\centering
\begin{tabular}{|c|c|c|c|c|c|c|}
\hline
\textbf{Noise} & \textbf{RL agent} & \textbf{Mean} & \textbf{SEM} & \textbf{25\%} & \textbf{50\%} & \textbf{75\%} \\
\hline
0.05 & Q & 8.37 & 0.04 & 8.15 & 8.30 & 8.50 \\
 & SF & \textbf{8.29} & 0.03 & 8.10 & \textbf{8.25} & 8.35 \\
 & Q($\lambda$ = 0.7) & 8.35 & 0.04 & 8.15 & 8.25 & 8.41 \\
 & Q($\lambda$ = 0.8) & 144.24 & 8.25 & 9.19 & 197.22 & 200.00 \\
 & Q($\lambda$ = 0.9) & 187.95 & 2.47 & 184.66 & 199.20 & 200.00 \\
 & PF($\lambda$ = 0.7) & 8.41 & 0.03 & 8.19 & 8.40 & 8.60 \\
 & PF($\lambda$ = 0.8) & 22.23 & 4.31 & 8.39 & 8.60 & 9.55 \\
 & PF($\lambda$ = 0.9) & 167.44 & 6.16 & 171.31 & 200.00 & 200.00 \\
\hline
0.25 &  Q & 75.27 & 2.34 & 55.64 & 72.72 & 91.67 \\
 & SF & 56.19 & 1.91 & 42.07 & 53.35 & 66.86 \\
 & Q($\lambda$ = 0.7) & 23.28 & 0.48 & 19.81 & 23.00 & 25.71 \\
 & Q($\lambda$ = 0.8) & \textbf{22.72} & 0.39 & 19.74 & \textbf{22.52} & 25.11 \\
 & Q($\lambda$ = 0.9) & 144.09 & 5.85 & 124.12 & 164.40 & 192.55 \\
 & PF($\lambda$ = 0.7) & 57.21 & 2.44 & 37.71 & 54.58 & 68.24 \\
 & PF($\lambda$ = 0.8) & 99.14 & 3.92 & 66.99 & 97.08 & 129.10 \\
 & PF($\lambda$ = 0.9) & 110.38 & 2.81 & 98.28 & 111.70 & 128.89 \\
 \hline
0.50 & Q & 98.90 & 1.73 & 86.83 & 97.80 & 110.67 \\
 & SF & 96.05 & 1.41 & 86.27 & 95.85 & 105.65 \\
 & Q($\lambda$ = 0.7) & 67.50 & 1.49 & 55.40 & 66.38 & 76.75 \\
 & Q($\lambda$ = 0.8) & \textbf{64.68} & 1.50 & 52.65 & \textbf{63.25} & 74.24 \\
 & Q($\lambda$ = 0.9) & 105.88 & 1.39 & 95.47 & 104.72 & 115.30 \\
 & PF($\lambda$ = 0.7) & 95.96 & 1.48 & 85.41 & 97.93 & 106.31 \\
 & PF($\lambda$ = 0.8) & 92.96 & 1.41 & 84.95 & 93.12 & 101.86 \\
 & PF($\lambda$ = 0.9) & 93.49 & 1.38 & 85.19 & 93.10 & 102.01 \\
 \hline
\end{tabular}
\caption{Distribution of Moving-averaged Episode Lengths in the Final Episodes Across Different Noise Levels in a Noisy 2D Environment. The table summarizes the episode length distributions for various RL agents in the final episodes, under noise levels $\sigma$ of 0.05, 0.25, and 0.50. The mean episode lengths, the standard error of the mean (SEM), and the quantiles (25\%, 50\%, 75\%) for each algorithm are shown.}
\label{tab:2D_episode_length}
\end{table}

\begin{figure}[h]
\centering
\includegraphics[width=1\textwidth]{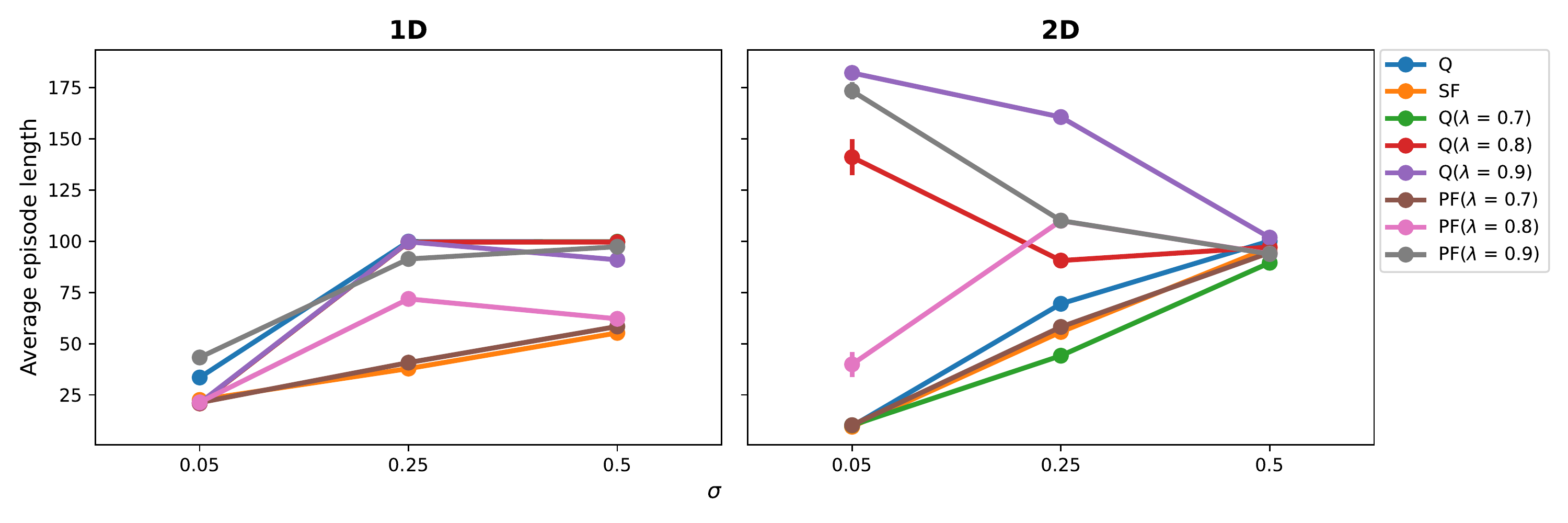}
\caption{Comparative Analysis of RL Algorithms in 1D and 2D Noisy Environments. The left panel depicts the overall average episode length in a 1D grid world, while the right panel shows the same in a 2D grid world, with both averaged across 100 trials. The y-axis indicates the mean episode length, representing the efficiency and consistency of learning for each algorithm. The noise levels, denoted by $\sigma$, are 0.05, 0.25, and 0.5, providing insight into each algorithm's robustness to environmental stochasticity and their adaptability across different complexities of spatial environments.}
\label{fig:noise_relationship}
\end{figure}

To demonstrate the variance in RL algorithm adaptation to noise across 1D and 2D grid worlds, we analyze the variation in average episode lengths in response to noise levels (Figure \ref{fig:noise_relationship}). Contrasting with the moving-averaged episode lengths presented in the final episodes in Figures \ref{fig:step_length}B and \ref{fig:2D_step_length}B and Tables \ref{tab:episode_length} and \ref{tab:2D_episode_length}, here, the average episode length is computed for each trial across total 3000 episodes and subsequently averaged across 100 trials to provide a comprehensive representation.

In the 1D domain, there is a clear trend where most algorithms exhibited elongated average episode lengths with increasing noise level (the left panel of Figure \ref{fig:noise_relationship}). Notably, the SF algorithm demonstrated a robustness, consistently keeping the average episode length shorter compared to other algorithms.

Transitioning to the 2D grid world, the algorithms' responses to noise became more intricate (the left panel of Figure \ref{fig:noise_relationship}). At low noise levels ($\sigma$ = 0.05), Q($\lambda$)-learning and PF($\lambda$)-learning algorithms with high $\lambda$ values were hampered in their learning progress. However, with noise increasing to $\sigma$ = 0.25, these same algorithms displayed surprising enhancements in performance. This counterintuitive phenomenon points to a non-linear relationship between noise levels and learning efficiency. In contrast, at the highest noise level examined ($\sigma$ = 0.5),  it was noted that all algorithms tended towards adopting a suboptimal policy, indicating that elevated noise levels do not facilitate learning but rather result in a plateau in performance.

\section{Discussion and conclusions} \label{sec:discussion}


Our comprehensive analysis of RL algorithms in 1D and 2D grid worlds highlights the complexities of algorithmic behavior in noisy environments. This section discusses the implications of our findings, the potential for algorithmic improvement, and the broader impact on the field of natural computing.

\paragraph{On adaptation in 1D noise environments}

Our investigation into RL algorithm performance across 1D and 2D environments under various noise levels has yielded insightful contrasts. In the 1D setting, agent performance consistently deteriorated as noise levels increased, showcasing a linear relationship between noise and efficiency. This trend suggests that, in simpler environments, increased noise straightforwardly hampers the agent's ability to navigate optimally.

The robustness of the SF algorithm in 1D environment can be attributed to its distinct approach of decomposing the value function into separate components for reward and state transition \citep{dayan1993,lee2023d,barreto2019,barreto2017,borsa2018}. In our study, only the state observations were subject to noise, leaving the reward vector unaffected. Q-learning, conversely, appeared more vulnerable in noisy conditions, as it relies on direct Q-value estimation from these noisy observations \citep{fox2015a, moreno2006}. It's important to note, however, that for effective policy making, SF learning also requires internal Q-value estimation from the reward vector and transition matrix, making it susceptible to the effects of noise, albeit to a lesser extent.

Agents employing eligibility traces, particularly PF learning with larger $\lambda$ values, demonstrated a general sensitivity to noise. This vulnerability is likely attributable to the way noisy observations significantly influence the state representation, affecting the computation of the trace over a longer historical window \citep{grzes2008}.

\paragraph{On adaptation in 2D noise environments}

The dynamics shifted when we extended our analysis to the 2D grid world. Here, we observed a departure from the linear relationship, with certain algorithms like Q-learning and SF-learning experiencing performance degradation as noise intensified. Interestingly, the Q($\lambda$) and PF($\lambda$) algorithms exhibit improvements as noise levels increase, particularly with a $\lambda$ value of 0.9. This enhancement in performance at medium noise levels ($\sigma$ = 0.25) suggests that noise-induced exploration may assist these algorithms in bypassing suboptimal local optima, leading to the development of potentially more effective policies. However, at the highest noise levels ($\sigma$ = 0.5), effective policy learning seems absent. All algorithms face challenges in identifying optimal paths amidst high noise, indicating that the advantageous impact of noise on navigation diminishes or turns counterproductive at these intensities.

\paragraph{Optimizing $\lambda$ for noise resilience in RL}

In our investigation, the examination of how RL algorithms equipped with eligibility tracing adapt to noise has unearthed significant findings regarding the influence of the $\lambda$ parameter. Notably, a $\lambda$ setting of 0.7 consistently outperformed higher settings of 0.9 or 0.8 across both 1D and 2D environments, with this effect being particularly evident under low noise conditions ($\sigma$ = 0.05) in 2D grid word. This suggests that higher $\lambda$ values, which influence a broader spectrum of preceding states, might introduce susceptibilities in environments characterized by noise. Essentially, elevated $\lambda$ values could intensify the effects of noisy eligibility traces, further compounding the detrimental impacts of observational noise.

Such insights hint at the potential optimality of moderate lambda parameters, even within the context of advanced deep learning methodologies. This notion is supported by the findings of Daly and Amato \citep{daley2018}, who reported that $\lambda$ values within the 0.6 to 0.7 range were most efficacious, albeit contingent upon the specific characteristics of the environment in question.

Therefore, our study underscores the imperative for continued empirical investigations to pinpoint the most effective $\lambda$ parameter, especially in relation to the degree of environmental noise. Undertaking such research is crucial not only for enhancing the efficacy and resilience of RL algorithms but also for broadening their practical deployment in real-world settings, where noise is an unavoidable element.

\paragraph{Neurobiological foundations of RL algorithms in noisy environments}

In our investigation, we aligned SR learning and its derivative, SF learning, with neurobiological mechanisms of decision-making and navigation. These models encapsulate the brain's capacity to forecast future state occupancies within spatial environments, resonating with biological plausibility \citep{stachenfeld2017, lee2020a}. PF learning, through its propagation of TD errors across an expanded set of preceding states, mirrors the brain's synaptic tagging process. This process involves marking synapses for long-term memory formation \citep{gerstner2018}, akin to the RL principle of eligibility traces that allocate credit to prior states based on temporal proximity.

Neuromodulatory systems like dopamine and norepinephrine, pivotal in RL, offer a biological foundation for PF learning. Dopamine's temporally distributed release in response to reward prediction errors hints at its role in conveying reward information across multiple prior states \citep{shindou2019,pan2005}, while norepinephrine enhances past event representations in neural circuits, potentially aiding TD error propagation to distant states \citep{hong2022}.

Furthermore, the integration of eligibility traces in PF learning enhances synaptic tagging, facilitating precise and resilient credit assignment. Despite evidence suggesting eligibility traces' superiority in enhancing SF learning \citep{pitis2018, bailey2022}, our findings indicate a susceptibility to noise in PF learning with high $\lambda$ values. This underscores the significance of calibrating the $\lambda$ parameter amidst noise, hinting at the necessity for a biologically consistent decay parameter in synaptic tagging for effective noise resilience.

Within cognitive map theory frameworks, SF learning adopts a forward-looking perspective, emphasizing next-state predictions, whereas PF learning retrospectively focuses on past state history. Recent studies attribute the orbitofrontal cortex's role in understanding prospective and retrospective continuities, with the hippocampus learning sequences using SR and PR concepts \citep{knamboodiri2021}. These insights affirm the importance of both perspectives for adept navigation and suggest that insights into neural underpinnings of cognitive map learning could pave the way for more sophisticated and robust RL algorithms.

\paragraph{Limitations of the study and potential future research directions}
Our research contributes important insights into the behavior of RL algorithms within noisy environments. However, it is imperative to recognize certain limitations inherent to our study design and methodology that highlight areas for further investigation. 

First, while grid-world simulations are instrumental for controlled analysis, they may not capture the full spectrum of challenges inherent in real-world scenarios. This limitation could restrict the extrapolation of our results to more intricate systems and settings. However, the simplicity of grid-world simulations is beneficial, as it isolates the stochasticity arising from observational noise, allowing for focused study on this aspect.

Second, our algorithmic comparison was somewhat narrow, concentrating on Q-learning, Q($\lambda$)-learning, and the SF and PF algorithms, excluding a range of newer or alternative approaches. Nevertheless, the selection of SF and PF algorithms was motivated by their relevance to neurobiological implication. It's important to clarify that the primary aim of this research was not the development of noise-resilient algorithms, but rather to understand the implications of these specific algorithms within the context of noise.

Addressing the highlighted limitations opens up several key avenues for future research:

The exploration of complex and dynamic environments represents a pivotal direction for advancing RL research. Advancing beyond simplistic grid world simulations to study environments with dynamic features, such as moving obstacles and variable rewards, could yield deeper insights into the real-world applicability of RL algorithms \citep{zhou2024}.

Moreover, the comprehensive evaluation of RL algorithms, particularly those harnessing state-of-the-art deep learning innovations, is essential for assessing their efficiency and robustness in noisy conditions \citep{sun2023a, park2023}. This would allow for a more thorough evaluation of algorithmic performance and resilience in the face of noise.

Investigating the neurobiological underpinnings of SF and PF learning algorithms offers another fruitful avenue for research. By exploring their connections to hippocampal function and cognitive mapping, researchers can deepen our understanding of the biological plausibility and efficiency of these algorithms \citep{fang2023, bono2023, george2023, ekman2023}.

The practical application of RL algorithms, especially for tasks like robotic navigation in dynamic environments, remains an area ripe for further investigation to ascertain their real-world efficacy. Conducting empirical tests of these algorithms in scenarios that mirror the unpredictability and complexity of actual conditions will be crucial for assessing their utility and adaptability \citep{ghadirzadeh2021, kaufmann2023, smith2023}.

Lastly, the significance of optimizing algorithmic parameters, such as the $\lambda$ values in eligibility trace-based algorithms, necessitates in-depth exploration. The pronounced effect of $\lambda$ settings on the performance of these algorithms underscores the need for systematic studies focused on fine-tuning parameters. The adoption of automated tuning methods, aimed at calibrating algorithm configurations for enhanced efficiency in diverse environments, represents a promising direction for enhancing the robustness and applicability of RL algorithms \citep{young2018}.

\paragraph{Conclusion}

This investigation elucidates the intricate, non-linear interplay between the sophistication of environmental settings and the efficacy of RL algorithms in contending with noisy conditions and executing resilient decision-making processes. Notably, algorithms that incorporate eligibility traces exhibit varied responses to environmental noise, influenced significantly by the adjustment of the $\lambda$ parameter within 2D environment. The observed resilience of SF algorithms, which may hint at connections to neurobiological processes, emerges as a significant insight from our study. In summation, venturing into the identified avenues of future research holds the promise of enriching our comprehension of RL algorithms and spearheading the creation of advanced AI systems adept at traversing the complexities and unpredictabilities inherent in real-world scenarios.

\backmatter

\bmhead{Acknowledgments}

This study was supported by the National Research Foundation of Korea(NRF) grant funded by the Korea government(MSIT; Ministry of Science and ICT)(No. NRF-2017R1C1B507279). The author would like to thank ChatGPT for their assistance in editing and improving the language of the paper, as well as for their helpful brainstorming sessions.

\section*{Statements and Declarations}

\begin{itemize}
\item Funding : This study was supported by the National Research Foundation of Korea(NRF) grant funded by the Korea government(MSIT; Ministry of Science and ICT)(No. NRF-2017R1C1B507279).
\item Conflict of interest/Competing interests : The author have no conflicts of interest to disclose.
\item Ethics approval : Not applicable
\item Consent to participate : Not applicable
\item Consent for publication : Not applicable
\item Availability of data and materials : Not applicable
\item Code availability : \href{https://github.com/HyunsuLee/PF_SF_noisy}{Github repository}
\item Authors' contributions : \textbf{Hyunsu Lee}: Conceptualization, Methodology, Software, Visualization, Writing, Funding acquisition.

\end{itemize}

\begin{appendices}

\section{Supplementary figures}\label{sec:appe_1}

\begin{figure}[H]
\centering
\includegraphics[width=1\textwidth]{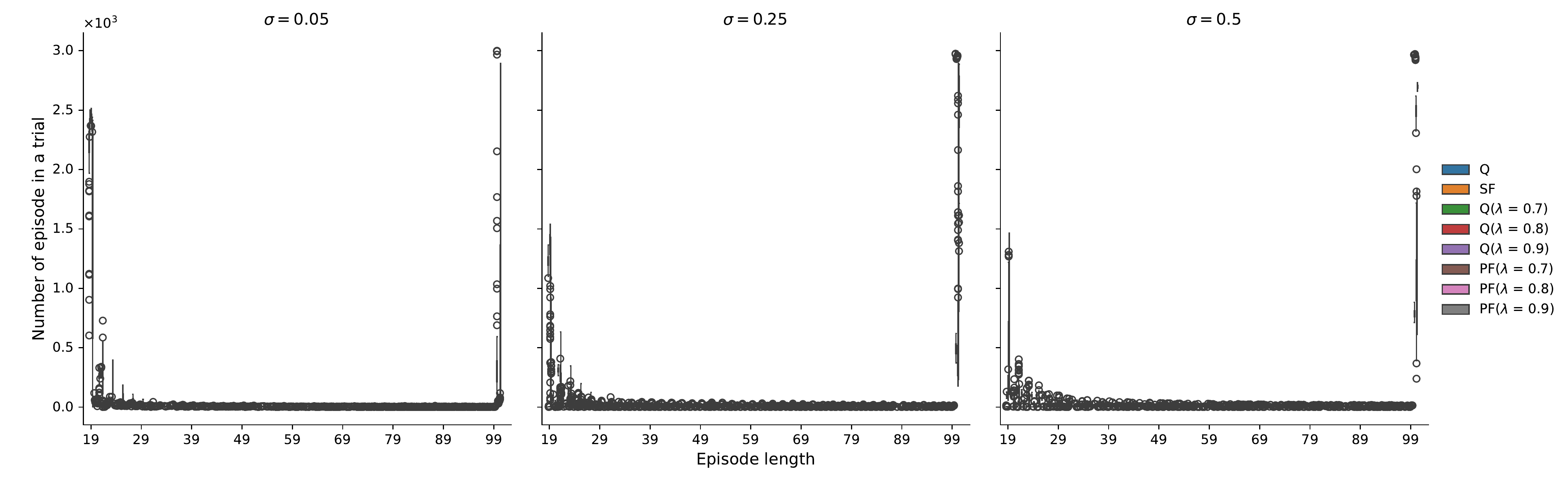}
\caption{Distribution of Episode Lengths for RL Algorithms at Different Noise Levels in a Noisy 1D Environment. This figure illustrates the frequency distribution of episode lengths across 3000 trials for various RL algorithms, categorized by noise levels $\sigma$ = 0.05, 0.25, and  0.5. Each plot corresponds to a different noise level, showcasing the number of episodes completed within a certain number of steps. Data points represent the episode count for a given length, providing a visual comparison of algorithm efficiency and consistency.}
\label{fig:whole_step_dist}
\end{figure}


\begin{figure}[H]
\centering
\includegraphics[width=1\textwidth]{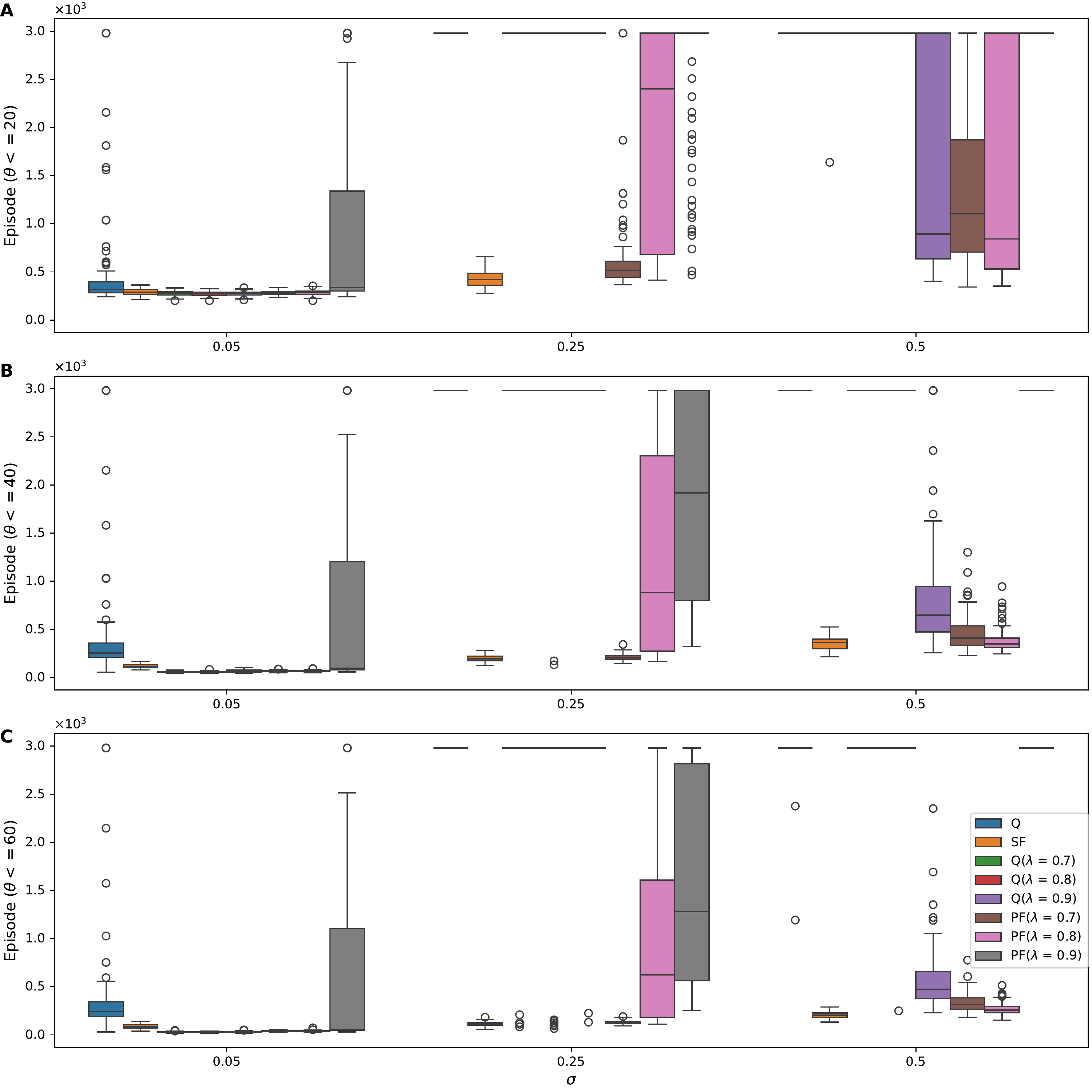}
\caption{Distribution of Episodes Reaching 20 (A), 40 (B), 60 (C) Steps or Fewer. The box plots detail the median, interquartile range, and outliers, offering insight into each algorithm's efficiency at each step threshold ($\theta$) under noise levels $\sigma$ = 0.05, 0.25, and  0.5.}
\label{fig:step_threshold_box}
\end{figure}

\end{appendices}

\bibliography{pr_noisy.bib}

\end{document}